\documentclass[10pt]{style}

\usepackage[numbers, square]{natbib}

\usepackage{graphicx}
\usepackage{dsfont}
\usepackage{braket}
\usepackage[english]{babel}
\usepackage{bbm}
\usepackage{nicefrac}
\usepackage{xfrac}
\usepackage{url}            % simple URL typesetting
\usepackage{booktabs}       % professional-quality tables
\usepackage{subcaption}
\usepackage{csquotes}
\usepackage{booktabs}
\usepackage{multirow}
\usepackage{tikz}
\usetikzlibrary{arrows.meta, positioning, shapes.geometric, calc, fit}
\usepackage[colorinlistoftodos,bordercolor=purple,backgroundcolor=pink!20,linecolor=pink,textsize=scriptsize]{todonotes}
\usepackage{wrapfig}

\usepackage{thm-restate}
\usepackage{empheq}

\renewcommand*{\backrefalt}[4]{%
    \ifcase #1 \footnotesize{(Not cited.)}%
    \or        \footnotesize{(Cited on page~#2)}%
    \else      \footnotesize{(Cited on pages~#2)}%
    \fi}

\usepackage{algorithm}
\usepackage{algpseudocode}
\usepackage[capitalize]{cleveref}
\usepackage[theorems,skins,most]{tcolorbox}
\theoremstyle{plain}

\theoremstyle{definition}

\theoremstyle{remark}

\crefname{assumption}{assumption}{assumptions}
\Crefname{assumption}{Assumption}{Assumptions}

\newcommand{\EE}{\mathbb{E}}

\newcommand{\R}{\mathbb{R}}
\newcommand{\N}{\mathbb{N}}

\newcommand{\1}{\mathds{1}}

\newcommand{\eqdef}{\coloneqq}

\def\<#1,#2>{\langle #1,#2\rangle}

\definecolor{lxs}{RGB}{138,43,226}

\NewDocumentCommand{\Var}{somo}{\mathrm{Var}\IfValueT{#2}{_{#2}}{} \IfBooleanTF{#1}{#3}{\IfValueTF{#4}{\!\left(#3\ \middle|\ #4\right)}{\parentheses*{#3}}}}

\newcommand{\bpi}{\pi_0}
\newcommand{\calset}[1]{\mathcal{#1}}

\newcommand{\data}{\calset{D}}
\newcommand{\Q}{Q^\pi}
\newcommand{\Qn}{Q^{\pi_n}}

\newcommand{\V}{V^\pi}

\newcommand{\A}{A^\pi}

\newcommand{\bQn}{\bar{Q}^{\pi_{n}}}

\title{What Matters for Simulation-to-Online Reinforcement Learning on Real Robots}

\reportnumber{} 

\author[1]{Yarden As}
\author[2]{Dhruva Tirumala}
\author[1]{René Zurbrügg}
\author[1]{Chenhao Li}
\author[1]{Stelian Coros}
\author[1]{Andreas Krause}
\author[2]{Markus Wulfmeier}

\affil[1]{ETH Zurich}
\affil[2]{Google DeepMind}

\correspondingauthor{yardas@ethz.ch, Code is publicly available at https://github.com/yardenas/panda-rl-kit}

\begin{abstract}
\textbf{Abstract:}
We investigate what specific design choices enable successful online reinforcement learning (RL) on physical robots. Across 100+ real-world training runs on three distinct robotic platforms, we systematically ablate algorithmic, systems, and experimental decisions that are typically left implicit in prior work. Rather than proposing a new algorithm, we study how standard off-policy RL can be made reliable in the sim-to-online setting. We find that some widely used defaults can be harmful, while a set of robust, readily adopted design choices within standard RL practice yields stable learning across tasks and hardware. These results provide the first large-sample empirical study of such design choices, enabling practitioners to deploy online RL with lower engineering effort.
\end{abstract}

\begin{document}
\maketitle

\section{Introduction}
We typically imagine learning agents as those capable of  learning continuously and adapting~``on-the-fly'' from information \emph{as it arrives} \citep{sutton,Hutter:24uaibook2}. Reinforcement learning (RL) formalizes how such agents can learn \emph{online} to act optimally within unknown dynamical systems. Despite the demonstrated success of RL in robotics \citep{hwangbo2019learning,tang2023industreal,chi2025diffusion,liao2025beyondmimic}, in most existing systems, learning occurs entirely offline, within a simulator or using a fixed dataset of demonstrations, leaving the canonical online learning setting far from being the standard practice. While this approach achieves decent success, without online adaptation, the resulting performance is ultimately limited by the quality of our priors. 

\looseness=-1What are the limits of relying on offline learning while avoiding firsthand real-world experience? Simulators are inevitably imperfect and the cost of obtaining high-quality, pre-training real-world data for robotics is orders of magnitude higher than in domains like language modeling---we do not have an Internet-scale corpus of real-world robotics data to scrape. This work is motivated by the realization that as tasks grow in complexity, future autonomous robotic systems must learn online, through embodied interaction, to continually adapt and improve competence in the open world \citep{javed2024the}.

\looseness=-1While running RL on real robots has been demonstrated in several works (e.g. \citet{haarnoja2018soft}, see \Cref{sec:related-works}), much of this literature focuses on demonstrating an idea within a relatively narrow real-world experimental setup. Additionally, prior work often focuses on a less realistic setting, where learning starts from scratch, which can lead to unsafe and wasteful exploration that could instead happen in simulation.

This paper takes a deliberately different perspective from much of the real-world RL
literature. We do not introduce a new RL algorithm but rather ask a pragmatic question:
given standard off-policy actor-critic methods, what combination of data and optimization choices makes simulation-to-online learning reliable on physical robots? Crucially,
because of the sim-to-real gap, naively fine-tuning a simulation-trained policy online can,
perhaps surprisingly, cause catastrophic forgetting during online training. We therefore focus on a controlled empirical study
across multiple robotic platforms, with the goal of understanding a simple recipe that practitioners
can directly adopt.

\paragraph{Our contribution.}
\looseness=-1
\begin{itemize}[leftmargin=0.5cm, parsep=0.3em]
\item We identify a simple recipe, built from standard off-policy RL components, that reliably
   fine-tunes simulation-trained policies online on real hardware: large-scale simulation
   pretraining, retained or warm-started replay, delayed actor updates with a conservative
   actor learning rate, and careful restoration of optimizer and target-network state.
\item \looseness=-1We validate this recipe through 100+ real-world training runs across three platforms:
   a vision-based Franka manipulation task, Unitree Go1 locomotion, and high-speed race-car
   navigation.
\item We systematically ablate the recipe ingredients and show that common defaults can fail:
   online-only replay, no warm start, and synchronous actor-critic updates can induce
   instability, while the complete recipe yields stable improvement.
\item We provide a two-timescale analysis that explains the failure mode of naive fine-tuning and why the recipe works: retained replay improves critic conditioning, while delayed actor updates reduce the perturbation caused by
   a critic that has not yet adapted to the online distribution.
\item We release an open-source implementation of our experiments. For the Franka Emika Panda robot, we provide a full-stack training pipeline that can be readily used to support ongoing RL research on real robots.
\end{itemize}
\begin{figure}
    \centering
    \vspace{12pt}

    % Subfigure 1
    \begin{subfigure}[t]{0.32\textwidth}
        \centering
        \begin{tikzpicture}
            \node[anchor=south west, inner sep=0] (image1) at (0, 0) {
                \includegraphics[width=\textwidth, trim={5cm 13cm 10cm 0}, clip]{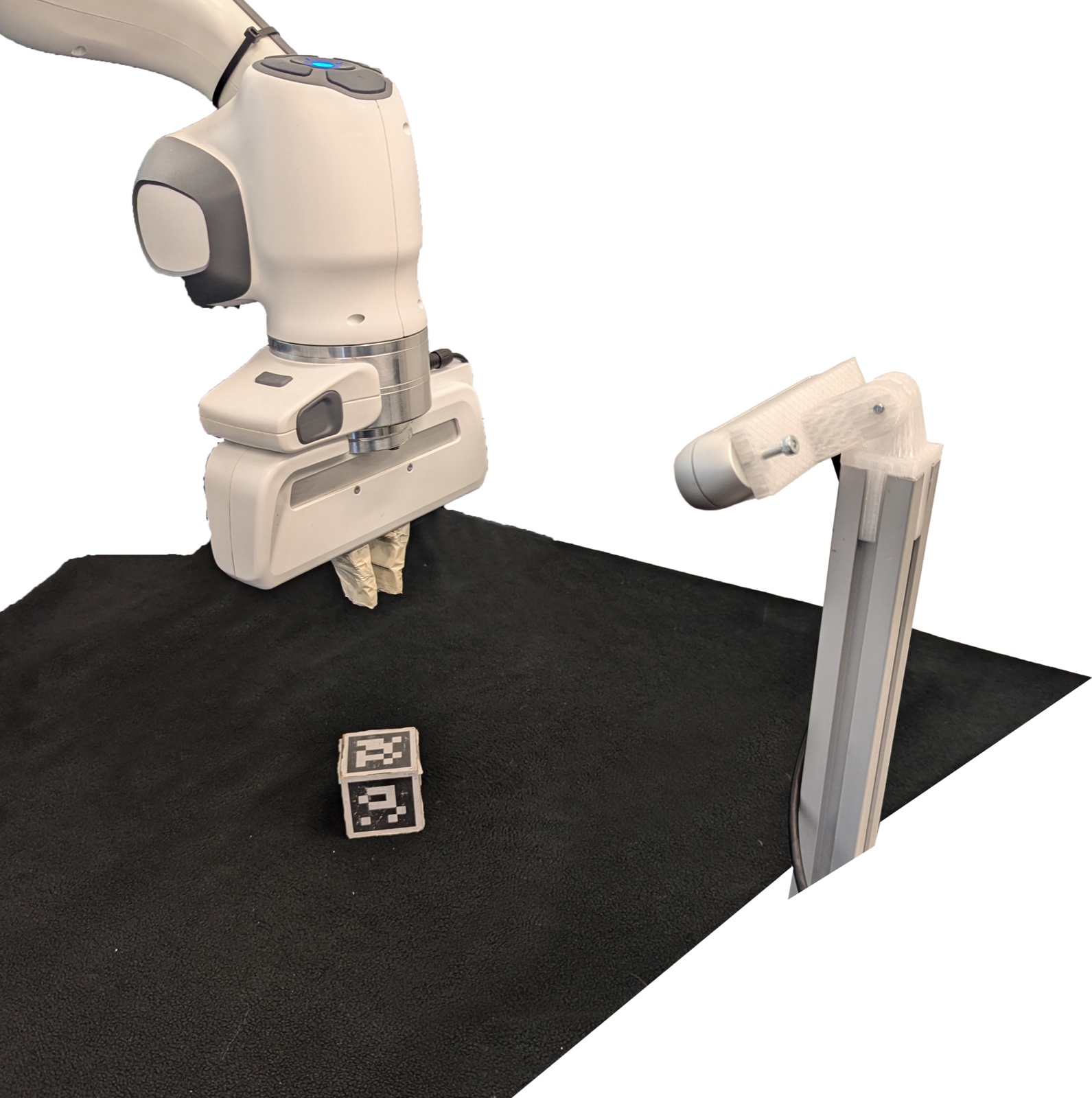}
            };

            % Obstacles label and curly arrows
            \node[anchor=south west, fill=white, font=\small, text=black, rounded corners, 
                xshift=-1.25cm, yshift=-0.75cm, align=center] (camera) at (image1.north east) {Realsense D455\\Camera};
            \draw[-{Circle[open]}, thick] (camera.south) to[out=-90, in=90] ++(-1.2, -1.35);
            % Goal label and curly white arrow (label on the right)
            \node[anchor=north west, fill=white, font=\small, text=black, rounded corners, 
                xshift=-2.25cm, yshift=-2.75cm, align=center] (goal) at (image1.north east) {Pick up\\cube};
            \draw[-{Circle[open, color=white]}, thick, white] 
                (goal.west) to[out=180, in=90] ++(-0.64, -0.34);
        \end{tikzpicture}
        \caption{Franka Emika Panda.}
        \label{fig:franka-demo}
    \end{subfigure}
    \hfill
    % Subfigure 2
    \begin{subfigure}[t]{0.32\textwidth}
    \centering
    \begin{tikzpicture}
        % Image of robot
        \node[anchor=south west, inner sep=0] (image2) at (0, 0) {
            \includegraphics[width=\textwidth]{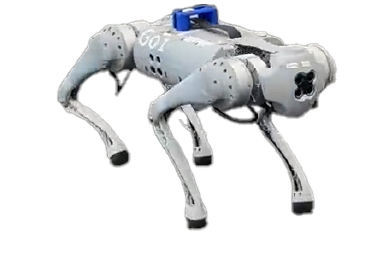}
        };
        % Define coordinates for annotation (tweak as needed)
        \begin{scope}[x={(image2.south east)},y={(image2.north west)}]
            \coordinate (center) at (0.5, 1.1);
            % Linear velocity arrow (forward direction)
            \draw[->, thick, black] ([shift={(0,0.0)}]center) -- ([shift={(0.26,-0.1)}]center)
                node[above right, black, xshift=-10pt, yshift=2pt] {$v$};
            
            \draw[->, thick, black, looseness=1.2]
                ([shift={(0,0.20)}]center) arc[start angle=90, end angle=420, radius=0.08]
                node[above left, black, font=\small, xshift=15pt] {$\omega_z$};
            
            \draw[->, thick, gray!70] 
                (center) -- ++(0, 0.3)
                node[above, black, font=\small] {$z$};
        \end{scope}
    \end{tikzpicture}
        \caption{Unitree Go1.}
        \label{fig:go1-demo}
    \end{subfigure}
    \hfill
    % Subfigure 3
    \begin{subfigure}[t]{0.32\textwidth}
    \centering
    \begin{tikzpicture}
        % Image of robot
        \node[anchor=south west, inner sep=0] (image2) at (0, 0) {
            \includegraphics[width=\textwidth, trim={0.85cm, 0.80cm, 2.15cm, 0.90cm}, clip]{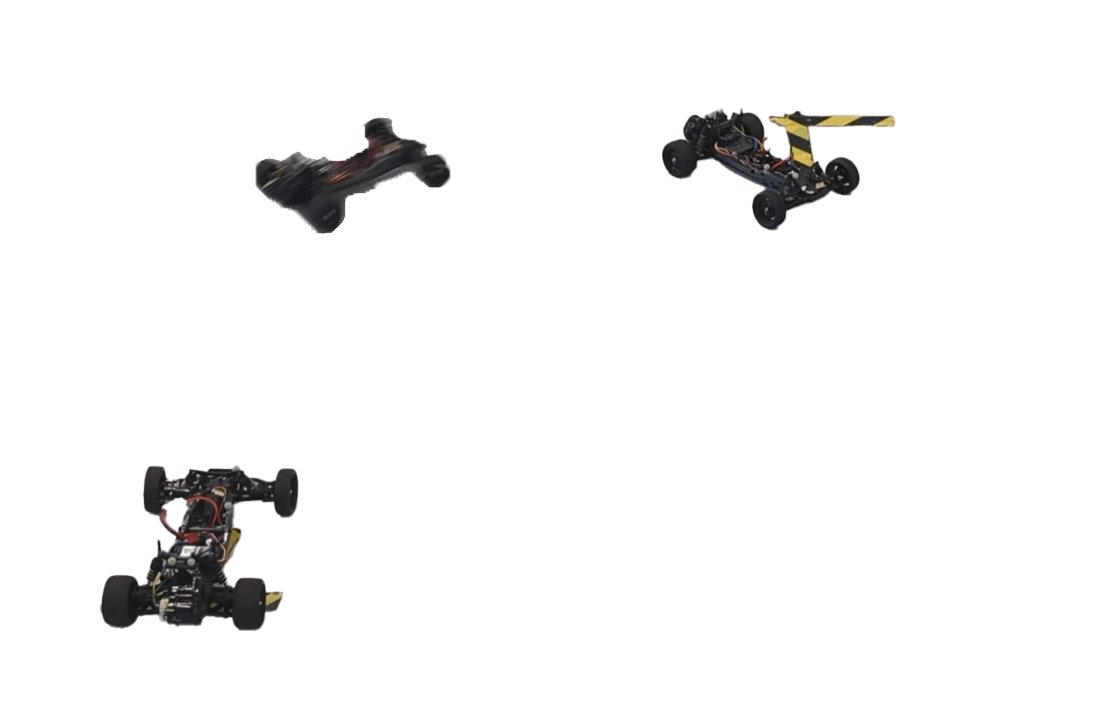}
        };
        % % Define coordinates for annotation (tweak as needed)
        \begin{scope}[x={(image2.south east)},y={(image2.north west)}]
            \node[anchor=north east, fill=white, font=\small, text=black, rounded corners, xshift=0.11cm, yshift=1.15cm] (goal) at (image2.north east) {Goal};
            \draw[-{Circle[open]}, thick] (goal.south) -- ++(0.0, -0.25);
            \node[anchor=north east, fill=white, font=\small, text=black, rounded corners, xshift=-2.75cm, yshift=1.15cm] (obstacles) at (image2.north east) {Race Car};
            \draw[-{Circle[open, color=black]}, thick] (obstacles.south) to[out=-90, in=90]  ++(0.1, -0.32);
            \draw[-{Circle[open, color=black]}, thick] (obstacles.east) to[out=-10, in=180]  ++(0.38, -0.31);
            \node[anchor=north east, fill=white, font=\small, text=black, rounded corners, xshift=-0.15cm, yshift=-1.5cm, text width=1.5cm, align=center] (start) at (image2.north east) {Start position};
            \draw[-{Circle[open, color=black]}, thick] (start.west) to[out=180, in=0]  ++(-0.38, -0.2);
        % --- Add dashed gray lines between car positions ---
            \coordinate (startCar) at (0.17, 0.35); % adjust as needed
            \coordinate (middleCar) at (0.24, 0.75); % Race Car position
            \coordinate (middleCarOut) at (0.44, 0.96); % Race Car position
            \coordinate (goalCar) at (0.72, 0.94); % Goal position
            \draw[dashed, thick, gray!70] (startCar) .. controls +(-0.0, 0.2) and +(-0.1, -0.15) ..  (middleCar);
            \draw[dashed, thick, gray!70] (middleCarOut) .. controls +(0.05, 0.05) and +(-0.0, 0.02) ..  (goalCar);
        \end{scope}
    \end{tikzpicture}
        \caption{Race Car.}
        \label{fig:rccar-demo}
    \end{subfigure}

    \caption{Robotic platforms studied in this work. We conduct our experiments on three robotic platforms spanning robotic manipulation, locomotion, and navigation tasks. \textbf{Manipulation.} We use a Franka Emika Panda robot to locate, grasp and lift a cube to a goal position. The policy determines the end-effector's position and the gripper's opening given grayscale image observations. \textbf{Locomotion.} We use a Unitree Go1 robot to follow joystick commands. The policy maps randomly sampled linear and angular velocity commands (expressed in the robot's local coordinate frame) to joint position targets. \textbf{Navigation.} Finally, we use a remote-controlled race car that must park at a specified goal position as quickly as possible. This task is particularly challenging due to the system's high agility, fast control loop (60~Hz) and the difficulty of accurately modeling tire friction and drifting behavior.
    }
    \vspace{-0.5cm}
    \label{fig:robots}
\end{figure}
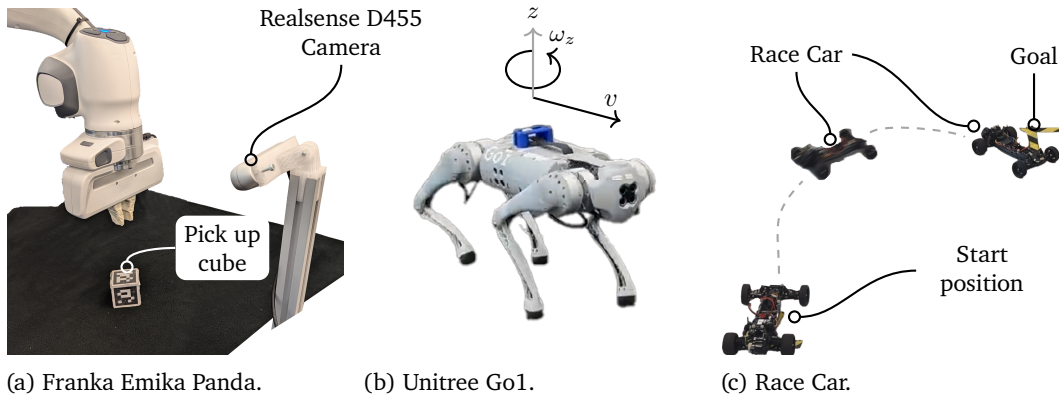

\section{Related Work}
\label{sec:related-works}
\looseness=-1
Early applications of RL on real robots date back to the 1990s (see \citealp{kober2013reinforcement} and references therein), focusing on relatively simple tasks with discretized, low-dimensional state-action spaces. Scaling RL to more challenging problems using ``deep RL'' techniques was achieved later by \citet{haarnoja2018soft}, who demonstrate that their algorithm, Soft Actor-Critic (SAC), learns efficiently on real-world robots. Since then, a few works use real-world robotic experiments to showcase the applicability of their algorithms in practice~\citep{ha2020learning,singh2020cog,nair2020awac,wu2023daydreamer,feng2023finetuning,zhou2025efficient,luo2025precise}. Many of these works rely on customized hardware or proprietary software, making them difficult to reproduce. Moreover, these works typically emphasize algorithmic innovation, without systematically examining the practical challenges of deploying RL on real-world robotic systems.

\looseness=-1In contrast, \citet{irbaz2021lessons} present a comprehensive review that encompasses several engineering and algorithmic challenges concerning deployment of RL on real robots. Similar to our work, they identify reusing data across trials and robots as key for improving sample efficiency and point to it as a promising direction for future work. However, they do not provide empirical evidence for the degree of sample efficiency improvements that can be gained by this approach. Further, \citet{tirumala2024replay} show the effectiveness of this idea across a wide range of simulated RL environments. We advance this line of work by demonstrating its effectiveness on real robots, where success cannot be achieved without high sample efficiency.

\looseness=-1Closely related to our work, \citet{yin2025sgft} study sim-to-real finetuning and report that SAC suffers from a performance drop upon transfer. They propose to fine-tune policies online based on reward shaping that drives the agent to state-actions in which the observed reward and the simulated one differ. In this work, we study instead how to mitigate such transfer challenges without additional algorithmic complexity. Similarly, \citet{smith2022legged,smith2022walk} demonstrate transfer from simulators, but restrict their experiments only to legged-locomotion tasks. We extend these results to a broader set of robots and systematically study how to effectively transfer such policies.

\section{Background}
\label{sec:background}
\subsection{Problem Setting}
\label{sec:problem-setting}
\paragraph{Markov decision processes.}
\looseness=-1We focus on infinite-horizon Markov decision processes~\citep[][MDP]{puterman2014markov}, defined by the tuple $(\calset{S},\calset{A},p,r,\gamma,\rho_0)$, where $\calset{S} \subset \R^{d_{\calset{S}}}$ and $\calset{A} \subset \R^{d_{\calset{A}}}$ are continuous state and action spaces, with states $s_t \in \calset{S}$ propagating through time $t = 0, \dots, \infty$ according to actions $a_t \in \calset{A}$ and unknown stochastic transition dynamics $p(s_{t + 1} | s_t, a_t)$. After each action the agent receives a reward $r_t \eqdef r(s_t, a_t),\;r : \calset{S} \times \calset{A} \to \R$ that is used to define the utility with respect to task completion. We consider the class of stationary stochastic policies $\pi(a_t | s_t) \in \Pi$ that map states to~(distributions over) actions.
The goal in MDPs is to find a policy $\pi^*$ that maximizes the accumulated sum of discounted rewards
\begin{align}
    \label{eq:mdp}
    \pi^* \in \arg\max_{\pi \in \Pi} J(\pi) \eqdef \EE_{\pi}\left[\sum_{t = 0}^\infty \gamma^t r(s_t, a_t)\right], \  s_0 \sim \rho_0(\cdot), ~a_t \sim \pi(\cdot | s_t), ~ s_{t + 1} \sim p(\cdot | s_t, a_t),
\end{align}
where $\gamma \in [0, 1)$ is a discounting factor, $\rho_0$ denotes a distribution over initial states and the expectation is with respect to sequences $s_0, a_0, s_1, a_1, s_2, \dots$ that follow a Markov chain induced by the policy $\pi$, the dynamics $p$ and the initial state distribution $\rho_0$. The following standard definitions for the value $\V$, action-value $\Q$ and advantage $\A$ functions will be useful in subsequent discussions:
\begin{align}
    \begin{aligned}
        &\V(s) \eqdef \EE_{\pi} \left[\sum_{t = 0}^{\infty} \gamma^t r(s_t, a_t) \bigg| s_0 = s\right], \quad 
        \Q(s, a) \eqdef \EE_{\pi} \left[\sum_{t = 0}^{\infty} \gamma^t r(s_t, a_t) \bigg| s_0 = s, a_0 = a\right] \\
        &\hspace{12em}\text{with}~a_t \sim \pi(\cdot|s_t)~\text{and}~s_{t + 1} \sim p(\cdot | s_t, a_t).
    \end{aligned}
\end{align}
Given full knowledge of the dynamics $p$ and reward $r$, one can solve \Cref{eq:mdp} via (approximate) \emph{planning} algorithms such as policy or value iteration \citep{puterman2014markov,betsekas}. However, in most relevant robotics problems, such access to the dynamics or reward is only limited, therefore requiring data-driven approaches like RL to solve \Cref{eq:mdp} in practice.

\paragraph{Episodic online learning.}
We consider learning in finite episodes. In this setting, in each episode $n = 1, \dots, N$, the agent executes a policy $\pi_n$ for $T$ time steps, after which the robot is manually reset to some state $s_0 \sim \rho_0(\cdot)$. The data from episode $n$ are collected as $\data_n \eqdef \{(s_t, a_t, s_{t + 1}, r_t)\}_{t = 0}^{T - 1}$ and aggregated in a ``replay buffer'' $\data_{\leq n} \eqdef \bigcup_{n' = 0}^n \data_{n'}$ \citep{Lin1992,mnih2013playing}. While data is collected in finite-length episodes, the agent optimizes the infinite-horizon discounted objective in \Cref{eq:mdp}.
This setting requires manual human resets. Fully autonomous learning from a \emph{single trajectory}, without manual resets, remains an active area of research, both theoretically and in practice~\citep{eysenbach2018leave,sharma2021autonomous,sharma2022autonomous,sharma2022state} and is left for future work.

\paragraph{Priors.}
\looseness=-1
We study a setting where prior knowledge of the task is available.
Good priors are crucial, since learning from scratch on the real system is likely to be highly unsafe and time-consuming.
When considering simulators, we assume that we have access to a dynamics model $p_0$ of the real dynamics $p$. We use $p_0$ to extract a prior policy $\bpi$. We will use $\data_0$ to denote the data that was generated \emph{in simulation} to obtain $\bpi$. Due to the sim-to-real gap, $\bpi$ is likely to perform suboptimally on the real system, and therefore requires additional real-world data to solve the task.

\subsection{Online Transfer With Off-policy RL}
\paragraph{Why off-policy RL?}
\looseness=-1
A successful pipeline for many robotic tasks combines massively parallel simulators \citep{zakka2025mujoco,mittal2025isaac} with domain randomization~\citep{tobin2017domainrandomizationtransferringdeep} and model-free on-policy methods such as PPO \citep{schulman2017proximal}. This approach often yields excellent performance \citep{hwangbo2019learning}, especially in locomotion tasks, where simulators are more accurate. 
However, simulators often struggle to accurately model contact-rich or vision-based tasks with complex scenes, thus making real-world adaptation essential. Since online training on real robots is constrained by real-time execution, sample efficiency is critical. On-policy methods use only data collected from the current policy and discard previous experience, resulting in limited sample efficiency that restricts their practicality in real world robotic settings.

\paragraph{Learning off-policy.} 
\looseness=-1
In contrast, off-policy algorithms \citep{lillicrap2015continuous,haarnoja2018soft,abdolmaleki2018maximum,fujimoto2018addressing,peng2019advantage} retain past data, and can even reuse data from other experiments with suboptimal hyperparameters \citep{tirumala2024replay}, often leading to an order-of-magnitude improvement in sample efficiency. While it is sometimes suggested that off-policy algorithms can be challenging to train effectively in massively parallel simulators~\citep{raffin2025isaacsim}, in \Cref{sec:method} and \Cref{sec:off-policy-in-sim} we show that this is nevertheless feasible. Off-policy algorithms operate in an approximate policy iteration scheme. A parameterized action-value function $\Qn_{\phi}$ that evaluates the policy after episode $n$ is learned by iteratively fitting it to an estimate of the (1-step) Bellman backup, minimizing the following loss: 
\begin{align}
\begin{aligned}
    \label{eq:policy-evaluation}
    &\ell_{\phi}(\phi;\theta) \eqdef \EE_{(s_t, a_t, s_{t + 1}, r_t) \sim \data_{\leq n }} \left[\frac{1}{2}\left(\Qn_\phi(s_{t}, a_{t}) - y
    \right)^2\right] \\
    &\text{where}~y = r_t + \gamma \bar{V}^{\pi_{n}}(s_{t + 1}) ~\text{and}~\bar{V}^{\pi_n}(s_{t + 1}) \approx \bQn(s_{t + 1}, a_{t + 1}),~a_{t + 1}\sim \pi_{n}(\cdot|s_{t + 1})
\end{aligned}
\end{align}
and $\bQn$ is a ``target network'' that tracks previous copies of $\Qn_{\phi}$, typically via Polyak averaging~\citep{lillicrap2015continuous} of parameters $\phi$:
\begin{equation}
    \label{eq:polyak}
    \phi^{\text{target}}_{k + 1} = (1 - \tau)\phi^{\text{target}}_{k} + \tau\phi_k,~k=0,\dots,K,
\end{equation}
whereby $\tau \in (0, 1)$ and $K \in \N$ is the number of critic updates per episode, while $k$ indexes those updates. In the policy improvement step, a parameterized policy is then extracted from $\Qn_\phi$ by ascending the action-value function
\begin{equation}
   \label{eq:policy-improvement} \ell_\theta(\theta;\phi) \eqdef - \EE_{s \sim \data_{\leq n},\,a \sim \pi_\theta(\cdot \mid s)} \left[ \Qn_\phi(s,a) \right], 
\end{equation}
which is often done by back-propagating gradients through $\Qn_\phi$ into the policy parameters~\citep{lillicrap2015continuous,haarnoja2018soft,fujimoto2018addressing} via the reparameterization trick \citep{kingma2015variational}. We omit entropy bonus terms in \Cref{eq:policy-evaluation,eq:policy-improvement} for notational clarity.

\paragraph{Efficient learning and approximation.}
\looseness=-1 Off-policy algorithms often define the critic ``update-to-data'' (UTD) ratio $\eta \eqdef \nicefrac{K}{T}$ as the number of critic gradient updates per real-world transition~\citep{janner2019trust,chen2021randomized}. If the actor is updated once every $M$ critic updates, its UTD is $\nicefrac{\lfloor K/M\rfloor}{T}$ (equal to $\eta/M$ when $M$ divides $K$). While increasing the critic UTD $\eta$ improves sample efficiency---crucial for fast online learning---it can also amplify approximation errors and overfitting~\citep{nauman2024overestimation}.
\citet{fujimoto2018addressing} mitigate this by interleaving the actor's update every $k = M, 2M, 3M, \dots, \lfloor K/M\rfloor M$ critic steps, where $M \in \N$. In this work we show that under abrupt changes in the dynamics, increasing $M$ is crucial for stable transfer.

\section{Unstable Transfer}

\paragraph{Distribution shifts and the ``downward spiral''.}
\looseness=-1
In fully online RL, policy-induced distribution shifts are often gradual because
the replay buffer is continually refreshed with current-policy data
\citep{mnih2013playing,lillicrap2015continuous,haarnoja2018soft,
fujimoto2018addressing}. In sim-to-online learning, however, deployment can
produce an abrupt shift. The pretrained policy $\pi_0$ selects actions that are favorable under the
simulator dynamics $p_0$, but its deployment trajectories are generated under
the real dynamics $p$. Consequently, the simulator-trained critic need not
remain accurate under deployment. Specifically, transitions alter future returns
even at familiar state-action pairs and may additionally carry the policy into states that were rarely encountered during pretraining.
The ``downward spiral''~\citep{nair2020awac,ball2023efficient,song2023hybrid,
nakamoto2024steering,zhou2025efficient} arises when the critic is inaccurate in
these shifted regions. Actions with overestimated values of $\Qn_\phi(s_t,a_t)$ are reinforced by the actor
update, causing the next policy to visit even more poorly covered states. This behavior is demonstrated empirically in \Cref{fig:stability-analysis}. 

\begin{figure}
    \centering
    \begin{tikzpicture}[
        node distance = 3.25cm,
        box/.style = {rectangle, draw, minimum width=2cm, minimum height=0.8cm, align=center},
        arrow/.style = {->, >=Stealth, very thick}
    ]
    % Main process nodes
    \node[label={[text width=4.5cm, align=center, xshift=-0.25cm]below:{State-action pairs with large errors $|\epsilon(s, a)|$ are added to $\data_{\leq n}$.}}] (collect) {Deploy $\pi_{n}$};
    \node[right=of collect, label={[text width=5cm, align=center]below:{\Cref{eq:policy-evaluation} is used to update $\Qn_{\phi}$; state-action pairs with large errors are sampled from $\data_{\leq n}$, leading to high returns being wrongly assigned to those state-action pairs \citep{fujimoto2018addressing}.}}, xshift=-0.5cm] (evaluate) {Evaluate $\pi_{n}$};
    \node[right=of evaluate, label={[text width=4.5cm, align=center]below:{Policy improvement finds a policy that maximizes a \emph{biased} action-value function $\Qn_\phi$.}}] (improve) {Improve $\pi_{n}$};
    
    % Top label box
    \node[box, above=0.3cm of evaluate, minimum width=5.5cm, xshift=-0.75cm, thick] (repeat) {Iterate for $n = 0, \dots, N - 1$ episodes};
    
    % Main flow arrows
    \draw[arrow] (collect) -- (evaluate);
    \draw[arrow] (evaluate) -- (improve);
    
    % Loop arrow from policy up to repeat box (top loop)
    \draw[arrow, rounded corners] 
        (improve.east) -- ++(0.7,0) |- (repeat.east);
    
    % Loop arrow from repeat box down to collect (left side)
    \draw[arrow, rounded corners] 
        (repeat.west) -- ++(-3.5,0) |- (collect.west);
    \end{tikzpicture}
    \caption{Off-policy algorithms may lose stability due to approximation errors in action-value functions, leading to unlearning of the prior policy $\pi_0$ during online learning.}
    % \vspace{-0.25cm}
    \label{fig:spiral}
\end{figure}
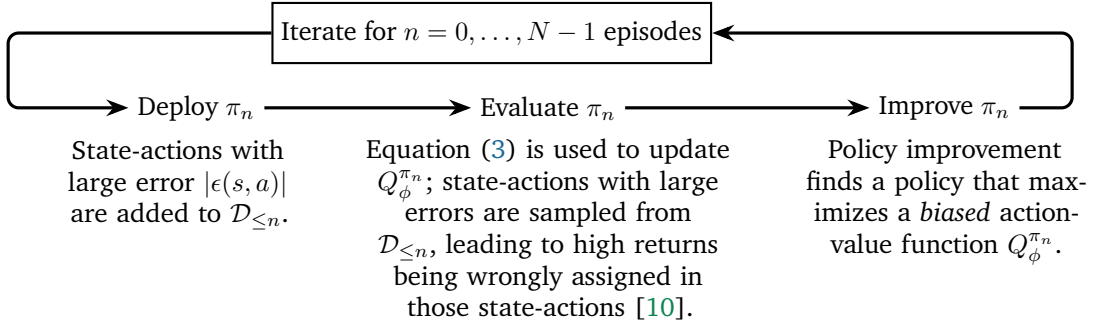

\begin{figure}
    \centering
    \includegraphics{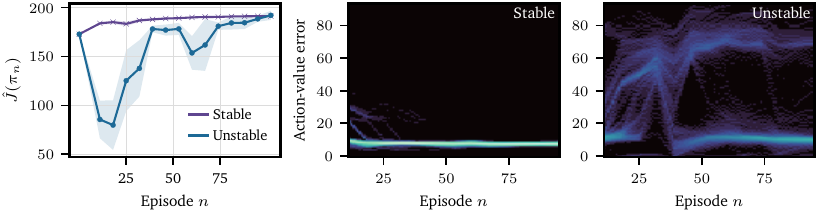}
    \caption{
    Downward spiral on a simulated Race Car robot under a mild dynamics
    mismatch.
    \textbf{Left:} Performance during online learning with vanilla Soft
    Actor-Critic (``Unstable'') and our approach (``Stable'').
    \textbf{Right:} Episode-wise histograms of the critic error
    $\epsilon(s_t,a_t)\approx
    \Qn_\phi(s_t,a_t)-\Qn_{\mathrm{MC}}(s_t,a_t)$ on newly collected data,
    with log counts represented by intensity. Vanilla SAC develops widespread
    positive errors as performance deteriorates, whereas our approach keeps
    most errors close to zero.
    }
    \label{fig:stability-analysis}
\end{figure}

\subsection{A Two-timescale View on the Downward Spiral}
To build understanding of the downward spiral, we view the critic as tracking
an equilibrium that moves as the policy and replay distribution change across
episodes. Following the two-timescale analysis of
\citet{konda2003onactor,konda2004convergence}, we study whether the critic can
adapt to this movement before the actor is updated.
Let $\nu_n$ denote the replay sampling distribution after episode $n$. During
that episode, $\pi_n$ interacts with the real dynamics $p$, and the collected
data shift the replay distribution from $\nu_{n-1}$ to $\nu_n$. Within the
$K$ critic updates following episode $n$, optimization blocks start at
$k=0,M,2M,\ldots,(\lfloor K/M\rfloor-1)M$. The block
starting at $k$ consists of critic updates $k$ through $k+M-1$, followed by
one actor update:
\begin{equation}
\begin{aligned}
\phi_{n,k+m+1}
&=
\phi_{n,k+m}
-
\beta_Q
\nabla_\phi
\ell_\phi(\phi_{n,k+m};\theta_{n,k},\nu_n),
\qquad m=0,\ldots,M-1,
\\
\theta_{n,k+M}
&=
\theta_{n,k}
-
\beta_\pi
\nabla_\theta
\ell_\theta(\theta_{n,k};\phi_{n,k+M},\nu_n).
\end{aligned}
\label{eq:two-timescale-updates}
\end{equation}
Thus, each episode contains $\lfloor K/M\rfloor$ complete blocks and
$K-M\lfloor K/M\rfloor$ remaining critic-only updates. The replay distribution
$\nu_n$ is fixed across all of these updates and changes only at episode
boundaries.
Let $\phi_{\mathrm{eq}}(\theta,\nu)$ denote a local critic equilibrium,
satisfying
\begin{equation}
\nabla_\phi
\ell_\phi
\bigl(
    \phi_{\mathrm{eq}}(\theta,\nu);
    \theta,\nu
\bigr)
=
0.
\end{equation}
At the beginning of the block starting at critic update $k$, the critic tracking
error is
\begin{equation}
e_{n,k}
\eqdef
\phi_{n,k}
-
\phi_{\mathrm{eq}}(\theta_{n,k},\nu_n).
\end{equation}
For the first block after an episode boundary, adding and subtracting the
equilibrium under the previous replay distribution gives
\begin{align}
e_{n,0}
&=
\underbrace{
\phi_{n,0}
-
\phi_{\mathrm{eq}}(\theta_{n,0},\nu_{n-1})
}_{\text{residual critic error}}
+
\underbrace{
\phi_{\mathrm{eq}}(\theta_{n,0},\nu_{n-1})
-
\phi_{\mathrm{eq}}(\theta_{n,0},\nu_n)
}_{\text{distribution-induced drift}}.
\label{eq:tracking-error-decomposition}
\end{align}
Thus, newly collected data can make a previously equilibrated critic
inaccurate at an episode boundary.
For subsequent blocks $k=M,\ldots,(\lfloor K/M\rfloor-1)M$, the analogous
decomposition has only the residual and actor-induced terms because $\nu_n$
remains fixed:
\begin{align}
e_{n,k}
&=
\underbrace{
\phi_{n,k}
-
\phi_{\mathrm{eq}}(\theta_{n,k-M},\nu_n)
}_{\text{residual critic error}}
+
\underbrace{
\phi_{\mathrm{eq}}(\theta_{n,k-M},\nu_n)
-
\phi_{\mathrm{eq}}(\theta_{n,k},\nu_n)
}_{\text{actor-induced drift}}.
\end{align}
Across the two cases, the critic equilibrium can therefore move because the
actor changes or because newly collected data shift the replay distribution.
Let
\begin{equation}
A_{n,k}
\eqdef
\nabla_{\phi\phi}^2
\ell_\phi
\bigl(
    \phi_{\mathrm{eq}}(\theta_{n,k},\nu_n);
    \theta_{n,k},\nu_n
\bigr),
\qquad
G_{n,k}
\eqdef
I-\beta_Q A_{n,k}.
\end{equation}
Assume that the local critic update is contractive, with
$q_{n,k}\eqdef\|G_{n,k}\|<1$. Under the corresponding first-order approximation, after
$M$ critic steps,
\begin{equation}
\phi_{n,k+M}
-
\phi_{\mathrm{eq}}(\theta_{n,k},\nu_n)
=
G_{n,k}^M e_{n,k},
\qquad
\left\|
G_{n,k}^M e_{n,k}
\right\|
\leq
q_{n,k}^M\|e_{n,k}\|.
\label{eq:critic-tracking}
\end{equation}

Define the actor update obtained with an equilibrated critic as
\begin{equation}
\theta_{n,k+M}^{\mathrm{eq}}
\eqdef
\theta_{n,k}
-
\beta_\pi
\nabla_\theta
\ell_\theta
\bigl(
    \theta_{n,k};
    \phi_{\mathrm{eq}}(\theta_{n,k},\nu_n),
    \nu_n
\bigr).
\end{equation}
Linearizing the actor gradient with respect to the critic parameters gives
\begin{equation}
\theta_{n,k+M}
-
\theta_{n,k+M}^{\mathrm{eq}}
=
-
\beta_\pi C_{n,k} G_{n,k}^M e_{n,k},
\qquad
C_{n,k}
\eqdef
\nabla_{\theta\phi}^2
\ell_\theta
\bigl(
    \theta_{n,k};
    \phi_{\mathrm{eq}}(\theta_{n,k},\nu_n),
    \nu_n
\bigr),
\end{equation}
and therefore
\begin{equation}
\left\|
\theta_{n,k+M}
-
\theta_{n,k+M}^{\mathrm{eq}}
\right\|
\leq
\beta_\pi
\|C_{n,k}\|\,
q_{n,k}^M
\|e_{n,k}\|.
\label{eq:actor-lag-bound-main}
\end{equation}

\looseness=-1
\Cref{eq:actor-lag-bound-main} captures the downward spiral. Dynamics mismatch when transitioning from the simulator to the real system initially moves the critic target away from the one learned in simulation,
producing a large tracking error $\|e_{n,k}\|$. After $M$ critic steps, the
residual error $G_{n,k}^M e_{n,k}$ distorts the actor update. The resulting policy
then changes the transitions collected in the next episode and hence moves
the critic target again. Critic error, policy change, and distribution shift
can therefore reinforce one another across episodes. Increasing $M$ reduces
the finite-tracking error through $q_{n,k}^M$, while decreasing $\beta_\pi$ limits
its effect on the actor.

\section{A Recipe for Stable Sim-to-Online RL}
\label{sec:method}
\label{sec:experiments}
For the rest of this work we focus on Soft Actor-Critic~\citep{haarnoja2018soft} as an off-policy algorithm as it delivers competitive results while maintaining overall robustness to hyperparameters. In \Cref{sec:additional-experiments} we present additional experiments with TD3 \citep{fujimoto2018addressing}.

\subsection{Recipe Overview and Experimental Setup}
\looseness=-1We validate the recipe on three real-world robots through controlled ablations after first establishing a reliable simulation-trained prior. Additional results and implementation details are provided in \Cref{sec:franka-details,sec:rccar-details,sec:go1-details,sec:off-policy-in-sim,sec:additional-experiments}.

\paragraph{Setup.}
Unless specified otherwise, in the following real-world experiments, we use Soft Actor-Critic \citep[SAC, ][]{haarnoja2018soft} together with the BRO  architecture for the critic $\Qn_\phi$ of \citet{nauman2024bigger}. We extend this architecture to vision control via DrQ \citep{yarats2021mastering} for the Franka Emika Panda robot.
We repeat each experiment with three random seeds and report the mean and standard error across seeds of each episode's undiscounted accumulated rewards, denoted by $\hat J(\pi)$. We use $T = 250$ for the Race Car and Franka Emika Panda and $T = 1000$ for the Unitree Go1.

\subsection{Simulation Pretraining}

\paragraph{Scaling Soft Actor-Critic.}
\looseness=-1
\begin{wrapfigure}[17]{r}{0.36\textwidth}
\vspace{-1.30\baselineskip}
\begin{center}
    \includegraphics{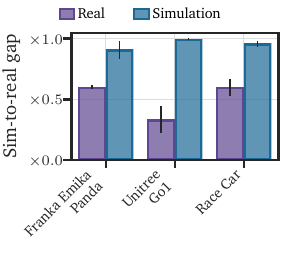}
\end{center}
\vspace{-0.5cm}
\caption{
\looseness=-1
Normalized performance w.r.t. the best experiment for each task across all experiments. In all tasks the prior policy suffers a performance decrease due to imperfect simulation dynamics.}
\label{fig:sim-to-real}
\end{wrapfigure}
Popular SAC implementations \citep[e.g.][]{brax2021github,huang2022cleanrl} perform one actor--critic update per parallel environment step, so the effective update-to-data ratio (UTD) $\eta$ decreases as the number of environments $N_e$ grows. While negligible for $N_e \sim 10$, this causes severe undertraining when $N_e \sim 1000$ needed for robust transfer (\Cref{sec:off-policy-in-sim}). We address this by increasing $\eta$ with $N_e$, though matching the data-generation rate exactly ($\eta \approx N_e$) is unnecessary due to diminishing returns. Sweeps over $\eta \in \{4, 8, 16, 32, 48, 64, 96, 128\}$ on the Franka Emika Panda ($N_e=512$) and Unitree Go1 ($N_e=8192$), each with five seeds, show that performance improves until a task-dependent saturation point, beyond which wall-clock cost rises substantially with little additional benefit (\Cref{fig:sweep-combined}).
\begin{figure}
    \centering
    % --- Top row: GO1 ---
    \begin{subfigure}[t]{0.49\textwidth}
        \centering
        \includegraphics{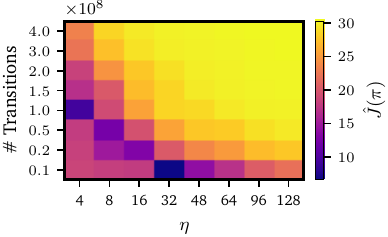}
        \label{fig:go1-sweep-performance}
    \end{subfigure}%
    \hfill
    \begin{subfigure}[t]{0.49\textwidth}
        \centering
        \includegraphics{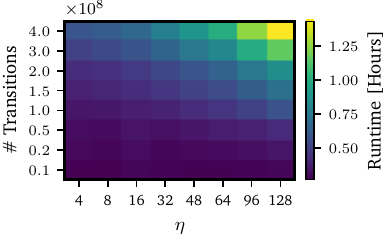}
        \label{fig:go1-sweep-runtime}
    \end{subfigure}
    \vspace{1em} % space between rows
    % --- Bottom row: Franka ---
    \begin{subfigure}[t]{0.49\textwidth}
        \centering
        \includegraphics{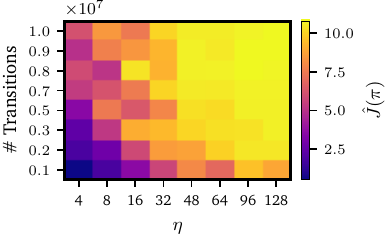}
        \label{fig:franka-sweep-performance}
    \end{subfigure}%
    \hfill
    \begin{subfigure}[t]{0.49\textwidth}
        \centering
        \includegraphics{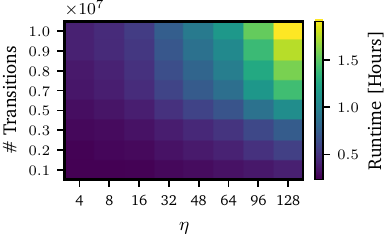}
        \label{fig:franka-sweep-runtime}
    \end{subfigure}
    \vspace{-0.5cm}
    \caption{
    Comparison of learning performance and runtime for different configurations.
    \textbf{Top:} Unitree Go1 experiments. \textbf{Bottom:}  Franka Emika Panda experiments.
    Increasing UTD requires fewer environment steps, but at the price of longer training time.
    }
    \vspace{-0.25cm}
    \label{fig:sweep-combined}
\end{figure}

\paragraph{Zero-shot performance.}
\looseness=-1
In the Franka Emika Panda setup (cf. \Cref{fig:robots}), $\pi_0$ successfully detects and approaches the cube, but it often fails to grasp and lift it on the real robot. These failures are primarily due to unmodeled contact dynamics between the gripper and the cube, as well as discrepancies between rendered and real visual observations.
For the quadruped, we train a constrained prior policy by limiting the range of commanded linear and angular velocities~($v$ and $\omega_z$ in \Cref{fig:go1-demo}) during simulation. While policies trained purely in simulation can already yield stable locomotion \citep[see][]{zakka2025mujoco}, limiting $\pi_0$ during pretraining allows us to demonstrate how online learning can be efficiently used in locomotion tasks.
In the Race Car environment, the resulting policy $\pi_0$ successfully parks the car in simulation; however, on the real system, it often overshoots the goal.
As shown in \Cref{fig:sim-to-real}, the prior policies succeed in solving all the simulated tasks, while exhibiting limited performance when transferring to the real system. Specific implementation details on each simulated task can be found in \Cref{sec:franka-details,sec:go1-details,sec:rccar-details}.

\subsection{Retained Replay}

\looseness=-1
By \Cref{eq:tracking-error-decomposition}, changes in the replay distribution
move the critic equilibrium and increase its tracking error. Through
\Cref{eq:actor-lag-bound-main}, this residual error distorts the next actor
update and can amplify the distribution shift in the following episode.
Retaining $\data_0$ anchors the replay distribution, causing the critic
equilibrium to move more gradually and thereby stabilizing early adaptation.
Accordingly, \citet{tirumala2024replay} and \citet{ball2023efficient} maintain
separate offline and online buffers,
$\data_0$ and
$\data_{\text{online}}\eqdef\data_{\leq n}\setminus\data_0$, and sample
\begin{equation}
    \label{eq:data-mixing}
    (s_t,a_t,s_{t+1},r_t)
    \sim
    (1-\alpha)\operatorname{Unif}(\data_0)
    +
    \alpha\operatorname{Unif}(\data_{\text{online}}),
    \qquad \alpha\in[0,1].
\end{equation}
They find that a balanced mixture ($\alpha=0.5$) accelerates offline-to-online
learning. In simulation-to-online learning, however, $\data_0$ reflects the
mismatched dynamics $p_0$; we therefore anneal $\alpha\to1$, using simulation
data to stabilize early adaptation while eventually learning entirely from
real-world transitions.

\begin{figure}
    \centering
        \centering
        \includegraphics{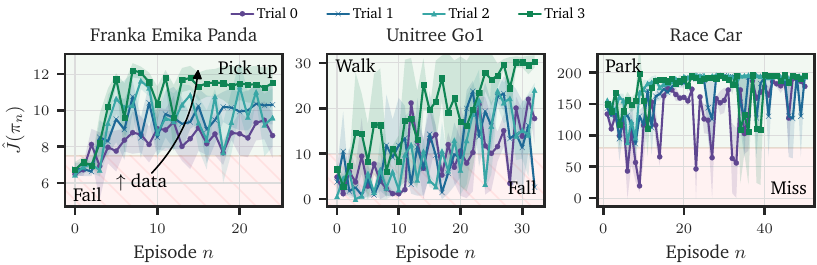}
        \caption{
        Performance and robustness increase as we accumulate more transitions over trials.
        Reusing training data across experiments accelerates online learning in all robots.}
        \vspace{-0.25cm}
        \label{fig:rae-all}
\end{figure}

\paragraph{Recycling data accelerates learning.}
\looseness=-1
We study the impact of retaining data from previous real-world experiments on learning performance. To show that, each experiment is composed of four trials that only share the same random seed. We run each experiment for three seeds for each robot. In the zeroth trial, learning is done only with online data collected in $\data_{\text{online}}$, which is saved at the end of each experiment. In subsequent trials, we load online replay buffers $\data_{\text{online}}$ of previous trials into $\data_0$ and start a fresh replay buffer $\data_{\text{online}}$. We then use \Cref{eq:data-mixing} to mix data between these two replay buffers, starting from $\alpha = 0.5$ and gradually increasing it to $\alpha = 1$ to reduce dependency on $\data_0$. \Cref{fig:rae-all} shows the performance increase as more data is retained. In \Cref{sec:additional-experiments} we ablate the initial choice of $\alpha$ and provide additional experiments showing improved stability when $\data_0$ is simulator data only.
As shown, across all tasks, significant performance gains can be achieved by retaining data from only a few preceding trials. \Cref{fig:franka-learning-demo} depicts trajectories before and after fine-tuning on the Franka Emika Panda. 

\subsection{Warm-Started Online Replay}
When $\data_0$ cannot be retained during online learning \citep{zhou2025efficient}, we instead approximate it by collecting data using the initial policy $\pi_0$ \emph{before} any updates to $\Qn_\phi$ or $\pi_n$. This warm-start collection is already standard in off-policy RL \citep[cf.][]{haarnoja2018soft}, and \citet{zhou2025efficient} show that it is crucial to mitigate instabilities in \emph{offline-to-online} RL. Practically, we simply run $\pi_0$ for $N^* \ll N$ trials~(without updating) and treat this as the initial buffer.

\begin{figure}
    \centering
        \centering
        \includegraphics{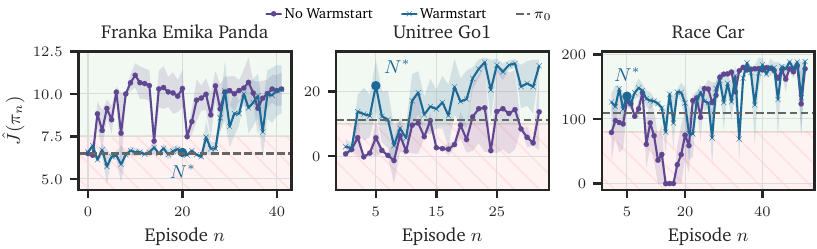}
        \caption{
        \looseness=-1The blue point marks the episode $N^*$ until which we do not update $\pi_0$ under the ``warmstart'' baseline.
        Prefilling $\data_{\text{online}}$ using $\pi_0$ improves learning stability in the Unitree Go1 and Race Car, while it is not required to obtain strong performance on the Franka Emika Panda.
        }
        \vspace{-0.25cm}
        \label{fig:wsrl-all}
\end{figure}

\paragraph{Warm starts approximate retained simulator replay.}
\looseness=-1We evaluate warm starts as an alternative for data retention. 
To this end, we do not load $\data_0$ but rather prefill $\data_{\text{online}}$ using a fixed copy of $\pi_0$ for $N^*$ iterations. For the Franka Emika Panda and Unitree Go1, we collect $5000$ transitions, corresponding to $N^* = 20$ and $N^* = 5$ respectively \citep[cf.][]{zhou2025efficient}. For the Race Car we use $1250$ transitions, which correspond to $N^* = 5$ episodes. The results are presented in \Cref{fig:wsrl-all}, where for the Franka Emika Panda robot, learning succeeds even without a warm start. In contrast, for the Unitree Go1 and Race Car robots, the performance drop is non-negligible.

\subsection{Delayed Actor Updates}

\begin{figure}
    \centering
        \centering
        \includegraphics{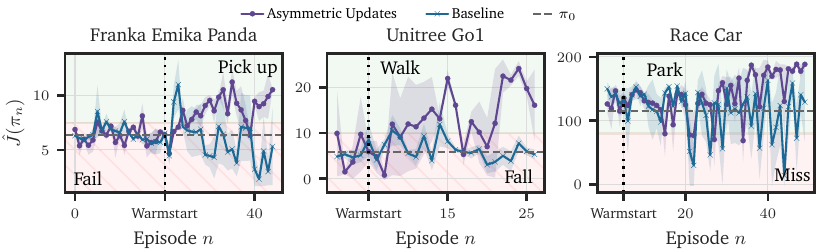}
        \caption{\looseness=-1We warmstart both runs with the same $N^*$, indicated by the vertical dotted line. Asymmetric updates are crucial for effective transfer across all robots, even when $\data_{\text{online}}$ is initialized via a warmstart.}
        \vspace{-0.45cm}
        \label{fig:asymmetric}
\end{figure}

\paragraph{Delayed actor updates stabilize fine-tuning.}
We analyze the importance of employing more conservative updates to the actor, interleaving its updates with more frequent critic updates while reducing its learning rate. Specifically, we update the actor every $20$ critic updates while reducing its learning rate (see \Cref{sec:implementation-details}) and compare this setup to a baseline that updates the actor every critic step, and uses a shared learning rate for the actor and critic. We present our results in \Cref{fig:asymmetric}, showing that for all robots, the baseline fails to improve performance due to training instability, while using asymmetric updates enables efficient transfer.
Notably, even when warm-starting $\data_{\text{online}}$ in these experiments, training without asymmetric actor-critic updates remains highly unstable.

\section{Conclusion}
\label{sec:conclusion}

\begin{figure}[H]
    \centering
    \begin{tikzpicture}[
        arrow/.style = {->, >=Stealth, thick}
    ]
        % Left part of the arrow
        \draw[-, thick] (1.5,0) -- (6.3,0);
        % Text in the middle (breaks the line)
        \node[below=-7pt] at (7.0,0.0) {Time};
        % Right part of the arrow
        % Add arrowhead at the end
        \draw[arrow] (7.7,0) -- (12.5,0);
    \end{tikzpicture}
    \begin{subfigure}[b]{\textwidth}
    \centering
    \begin{tabular}{c @{\hskip 5pt} c @{\hskip 5pt} c @{\hskip 5pt} c @{\hskip 5pt} c @{\hskip 5pt} c @{\hskip 5pt} c}
    \rotatebox{90}{$\pi_0$} &
        \begin{tikzpicture}
            \node[anchor=south west, inner sep=0] (image1) at (0, 0) 
                {\includegraphics[width=0.15\textwidth, clip]{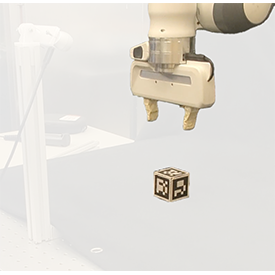}};
        \end{tikzpicture} &
        \begin{tikzpicture}
            \node[anchor=south west, inner sep=0] (image2) at (0, 0) 
                {\includegraphics[width=0.15\textwidth, clip]{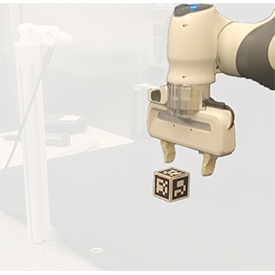}};
        \end{tikzpicture} &
        \begin{tikzpicture}
            \node[anchor=south west, inner sep=0] (image3) at (0, 0) 
                {\includegraphics[width=0.15\textwidth, clip]{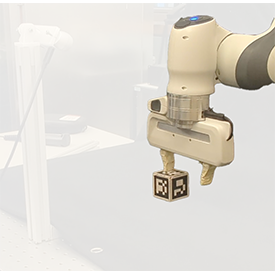}};
        \end{tikzpicture} &
        \begin{tikzpicture}
            \node[anchor=south west, inner sep=0] (image4) at (0, 0) 
                {\includegraphics[width=0.15\textwidth, clip]{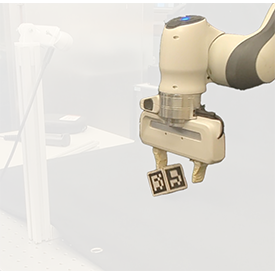}};
        \end{tikzpicture} &
        \begin{tikzpicture}
            \node[anchor=south west, inner sep=0] (image5) at (0, 0) 
                {\includegraphics[width=0.15\textwidth, clip]{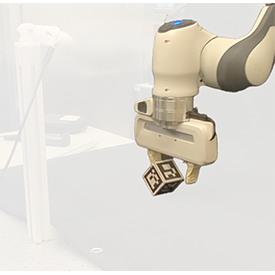}};
        \end{tikzpicture} &
        \begin{tikzpicture}
            \node[anchor=south west, inner sep=0] (image6) at (0, 0) 
                {\includegraphics[width=0.15\textwidth, clip]{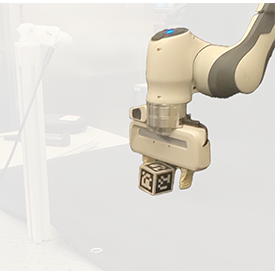}};
        \end{tikzpicture} \\
    \rotatebox{90}{$\pi_N$} &
        \begin{tikzpicture}
            \node[anchor=south west, inner sep=0] (image7) at (0, 0) 
                {\includegraphics[width=0.15\textwidth, clip]{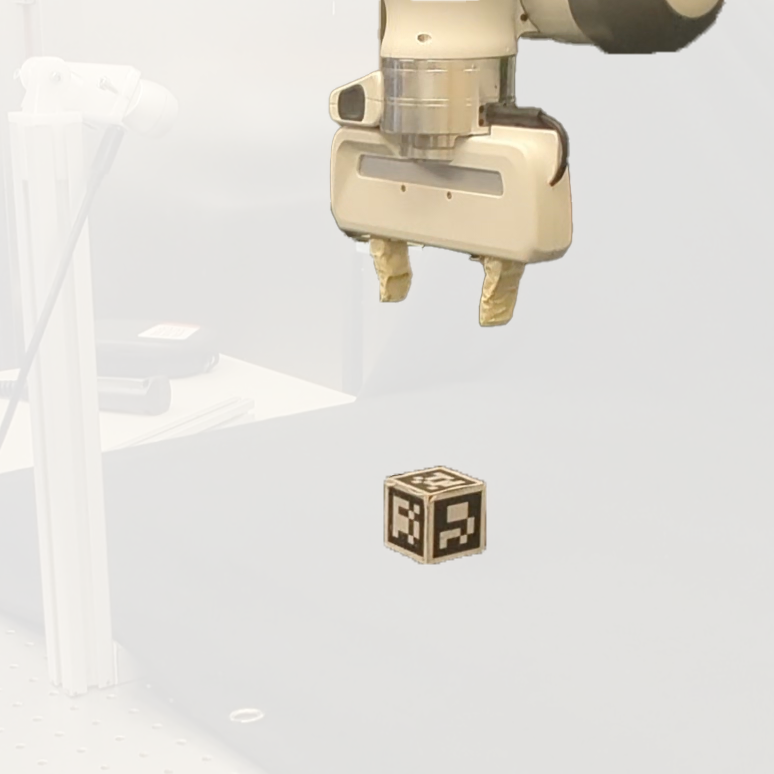}};
        \end{tikzpicture} &
        \begin{tikzpicture}
            \node[anchor=south west, inner sep=0] (image8) at (0, 0) 
                {\includegraphics[width=0.15\textwidth, clip]{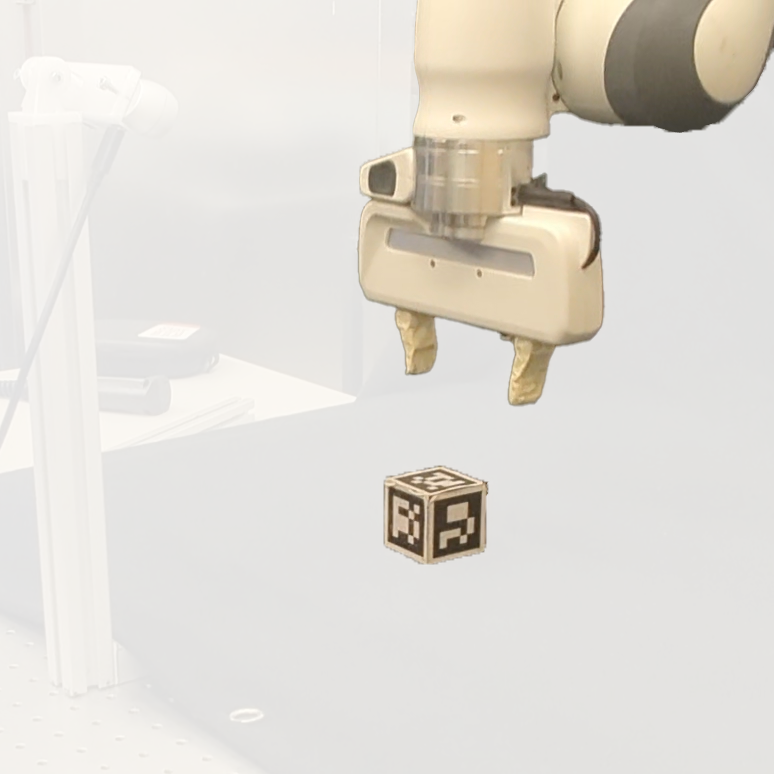}};
        \end{tikzpicture} &
        \begin{tikzpicture}
            \node[anchor=south west, inner sep=0] (image9) at (0, 0) 
                {\includegraphics[width=0.15\textwidth, clip]{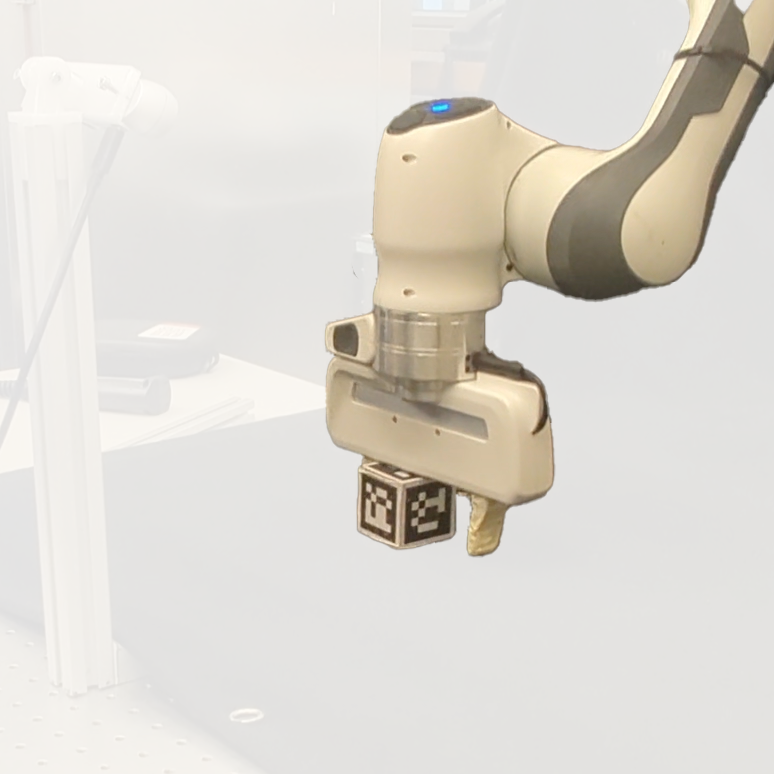}};
        \end{tikzpicture} &
        \begin{tikzpicture}
            \node[anchor=south west, inner sep=0] (image10) at (0, 0) 
                {\includegraphics[width=0.15\textwidth, clip]{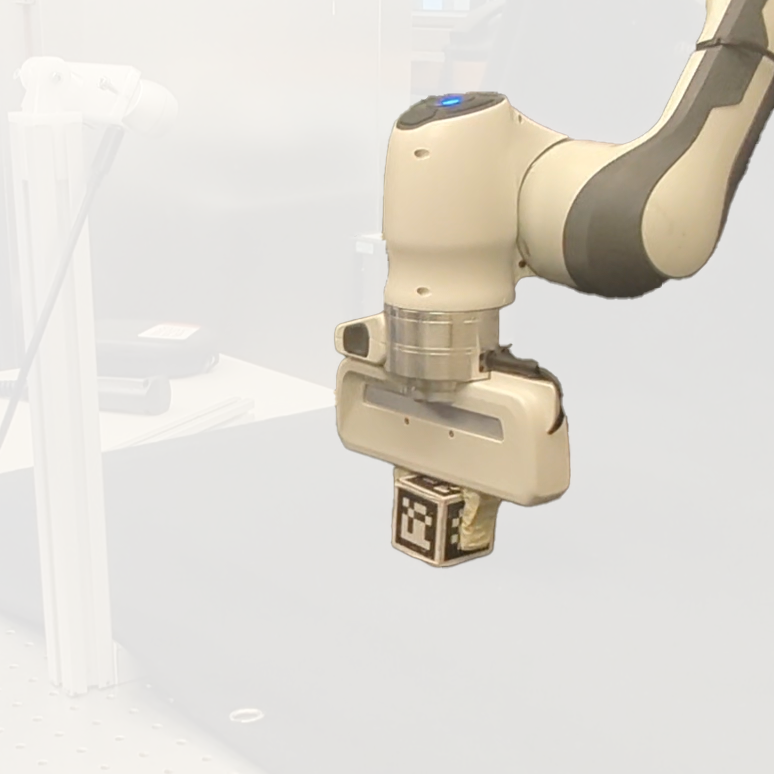}};
        \end{tikzpicture} &
        \begin{tikzpicture}
            \node[anchor=south west, inner sep=0] (image11) at (0, 0) 
                {\includegraphics[width=0.15\textwidth, clip]{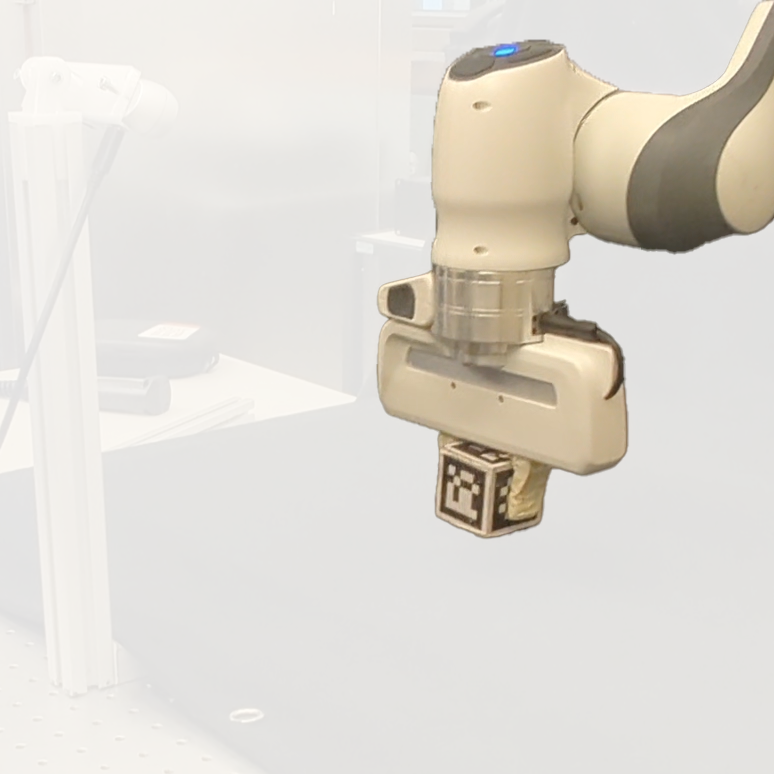}};
        \end{tikzpicture} &
        \begin{tikzpicture}
            \node[anchor=south west, inner sep=0] (image12) at (0, 0) 
                {\includegraphics[width=0.15\textwidth, clip]{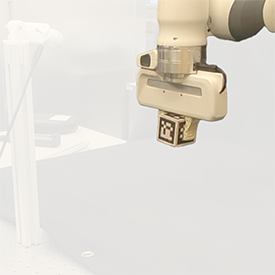}};
        \end{tikzpicture} \\
    \end{tabular}
    \end{subfigure}
    \caption{After roughly ten minutes of training, including additional hardware overhead (e.g., resetting the robot and data transmission over the internet), SAC recovers a policy with an almost-perfect success rate for the Franka Emika Panda robot.}
    \vspace{-0.50cm}
    \label{fig:franka-learning-demo}
\end{figure}

\looseness=-1
We present a large-scale empirical study of finetuning simulation-trained RL priors directly on hardware across three robotic platforms. Taken together, our experiments show that the recipe components enable standard SAC to improve simulation-trained priors across manipulation, locomotion, and navigation (\Cref{fig:rae-all,fig:wsrl-all,fig:asymmetric}). The resulting policies recover reliable grasping on the Franka Emika Panda, improve robustness to previously unseen commands on the Unitree Go1, and park the Race Car faster and more precisely, as shown in \Cref{fig:franka-learning-demo,fig:go1-learning-demo,fig:rccar-learning-demo}.

\looseness=-1
Based on these results, we provide guidance to make online RL on hardware more accessible to RL researchers and practitioners. Our experiments show that, despite training instabilities arising from distribution shifts, standard off-policy algorithms require no major modifications, and remain effective for fine-tuning policies within realistic time budgets, even for vision-based tasks with sparse rewards. Our results further highlight the \emph{opportunity} that lies in reusing data to efficiently scale to more complex tasks. While these findings advance the goal of making online RL more practical, they also raise several important research questions: How can we optimally select samples from offline data $\mathcal{D}_0$ to improve online sample efficiency? How can data be effectively reused across different tasks? Are there better regularization strategies that enable faster learning?
Finally, our work focuses on the semi-automated \emph{episodic} setting, where human intervention is still required for resets and safety. Developing practical algorithmic solutions that enable fully autonomous learning is a promising direction for future work.

\section*{Acknowledgements}
Y.A. received funding from grant no. 21039 of the Hasler Foundation and the ETH AI Center.
M.W. and D.T. participated in an advisory capacity in this project. We thank Manish Prajapat, Manuel Wendl, Dongho Kang and Mert Albaba for their advice on early revisions of this work. \Cref{fig:spiral} is inspired by Figure 12.3 in \citet{Hutter:24uaibook2}. 
The authors would like to thank the Robotics Systems Lab at ETH Zurich for providing access to the Franka Emika Panda robot.

\bibliography{references}
\bibliographystyle{unsrtnat}

\appendix
\section*{Appendix}
\section{Off-Policy Training in Massively-Parallel Simulators} 
\label{sec:off-policy-in-sim}
\begin{wrapfigure}[16]{r}{0.36\textwidth}
\vspace{-1.0\baselineskip}
  \begin{center}
        \includegraphics{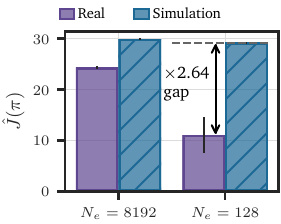}
  \end{center}
  \caption{Performance on the real robot after training with 8192 and 128 domain-randomized environments. A sufficient number of domain-randomized environments is key to effective transfer.}
  \label{fig:go1-rewards}
\end{wrapfigure}
Many off-policy algorithms were initially designed for the setting where the agent collects trajectories \emph{sequentially} in a \emph{single environment}. Despite that, a major advancement of RL in robotics leverages the ability to roll out thousands of simulated trajectories \emph{in parallel} to accelerate training. In fact, when combined with domain randomization, this approach is key to robust sim-to-real transfer, especially in locomotion tasks \citep{hwangbo2019learning}. 
However, this success primarily relies on on-policy algorithms, thus limiting its applicability to problems that can only be solved without additional online interaction. Off-policy methods, while more sample-efficient, require subtle yet nontrivial modifications to scale effectively in parallel simulation~\citep{raffin2025isaacsim}. Although prior works propose specialized algorithmic solutions for this setting \citep{li2023parallel,seo2025fasttd3}, we show that SAC remains effective with minimal modifications, enabling unified transfer from large-scale simulation to real-world fine-tuning.

\paragraph{Scale matters.}
\looseness=-1We show that using too few domain-randomized environments, denoted by $N_e$, leads to poor transfer to the real robot, even when SAC converges to a seemingly good policy in simulation. To demonstrate this, we train policies for the Unitree Go1 robot in simulation with $N_e \in \{128, 8192\}$. As shown in \Cref{fig:go1-rewards}, both configurations achieve similar performance in simulation. However, when deploying the policy trained with $N_e = 128$, the robot exhibits reduced stability and achieves significantly lower real-world rewards. This result shows that a large number of domain-randomized environments ($N_e \sim 10^3$) is essential for robust sim-to-real transfer. The implication of this result is that using a setting that is closer to ``vanilla'' SAC, with $N_e \sim 10$, is not sufficient for robust transfer to a real robot.
Note that for the sake of this demonstration, we do not limit the commands during pretraining as done in \Cref{sec:experiments}, meaning that the policy trained with $N_e = 8192$ transfers well without additional online training.

\section{More Experiments}
\label{sec:additional-experiments}

\paragraph{Zero-shot performance compared to PPO.}
Our work focuses mainly on off-policy algorithms due to their improved sample efficiency when training online. We validate that the drop in zero-shot deployment performance  on the real system with SAC is indeed due to the sim-to-real gap and not because of our choice of algorithm. 

\begin{wrapfigure}[14]{r}{0.36\textwidth}
\vspace{-1.25\baselineskip}
  \begin{center}
        \includegraphics{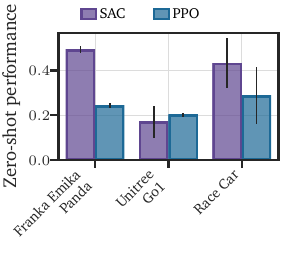}
  \end{center}
  \vspace{-1.25\baselineskip}
  \caption{
  Normalized performance of PPO and SAC w.r.t. the best performance achieved after online learning.}
  \label{fig:ppo-vs-sac}
\end{wrapfigure}
In particular, in \Cref{fig:ppo-vs-sac}, we plot the zero-shot performance of PPO on the same tasks. We focus on PPO as it is frequently used for sim-to-real deployment without additional training. In addition, to make a fair comparison, we use policy networks of the same size. For the Franka Emika Panda and Unitree Go1 robots, we use the same hyperparameters described by \citet{zakka2025mujoco}. We also note that for all environments, in simulation, SAC and PPO reach the same performance. As shown in \Cref{fig:ppo-vs-sac}, in all tasks the zero-shot performance of both SAC and PPO is significantly lower than the downstream performance after additional online learning with SAC.

\begin{figure}
    \centering
        \centering
        \includegraphics[width=\textwidth]{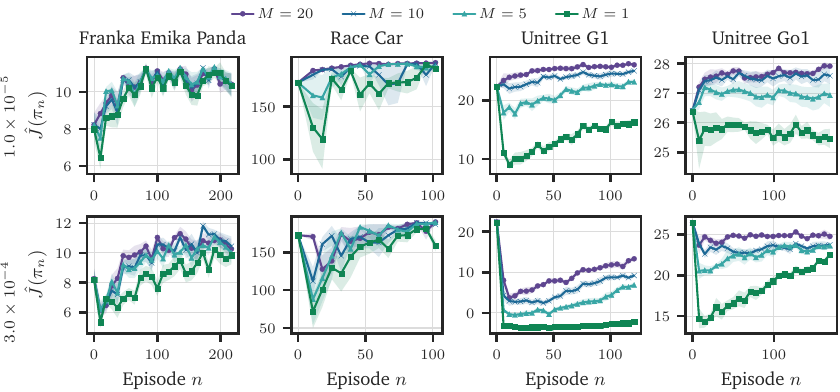}
        \caption{
        Learning curves for Soft Actor-Critic under mismatch in the dynamics. In the Franka Emika Panda robot, we replace the cube with a soft red ball. For the Race Car, we first pretrain on a semi-kinematic bicycle model and finetune on more realistic dynamics that account for tire friction \citep[see][]{kabzan2020amz}. In the Unitree G1 \citep[humanoid, see][]{zakka2025mujoco} and Go1 robots, we reduce the ground friction. We ablate $M \in \{20, 10, 5, 1\}$, showing significant stability improvements as we increase $M$ and reduce the learning rate from $3\times10^{-4} \rightarrow 1\times10^{-5}$ across all robots. In \Cref{sec:additional-experiments} we show similar results using TD3 \citep{fujimoto2018addressing}.
        }
        \vspace{-0.25cm}
        \label{fig:stability}
\end{figure}

\paragraph{Sim-to-sim ablation of $\boldsymbol{M}$.}
In \Cref{fig:stability}, we ablate different choices of $M$ and the actor's learning rate in a sim-to-sim transfer setup. As shown, increasing $M$, which corresponds to updating the actor less frequently, drastically improves learning stability across all robots. 

\begin{figure}
    \centering
        \centering
        \includegraphics[width=\textwidth]{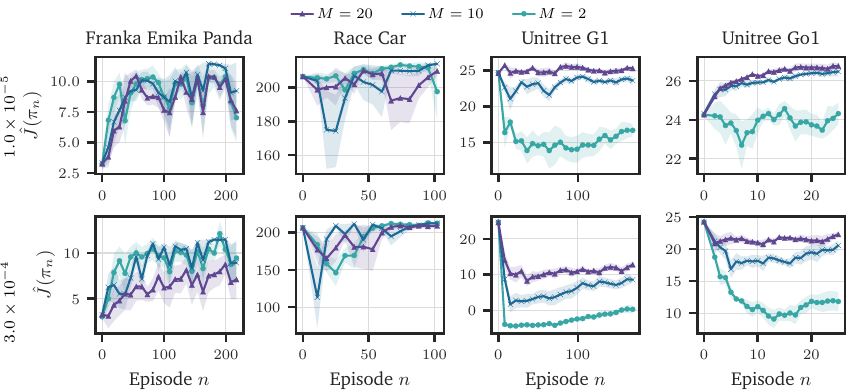}
        \caption{
        Learning curves for transfer (in simulation) of TD3. Similarly to the results in \Cref{fig:stability}, for small values of delay, mismatch in the dynamics can result in significant performance drops.
        }
        \label{fig:stability-td3}
\end{figure}

\paragraph{Sim-to-sim with TD3.}
We provide additional experiments with TD3 \citep{fujimoto2018addressing}, a state-of-the-art off-policy RL algorithm. TD3 delays policy updates by default; $M = 2$ is the default hyperparameter. Below, we repeat our sim-to-sim experiment in \Cref{fig:stability} but replace SAC with TD3. We present our results in \Cref{fig:stability-td3}, showing that it exhibits similar transfer dynamics to SAC during online learning.

\begin{figure}[h]
    \centering
    \includegraphics{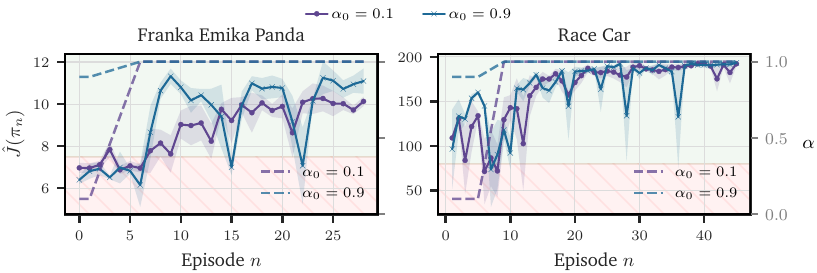}
    \caption{Dashed lines represent the value of $\alpha$ during learning.
    Robust transfer under different values of the initial mixing parameter $\alpha$.}
    \label{fig:alpha-ablation}
\end{figure}

\paragraph{Initial mixing $\boldsymbol{\alpha}$.}
We evaluate the choice of the initial value of $\alpha$ on learning stability and performance. We train with $\alpha_0 \in \{0.9, 0.1\}$ on the Franka Emika Panda and Race Car robots and report the results in \Cref{fig:alpha-ablation}. Our experiments show that as long as offline data is used at the onset of training, while online data dominates in the later part of training, good performance can be attained. As expected, using online data earlier in training leads to better performance while compromising training stability.

\begin{figure}[h]
    \centering
    \includegraphics{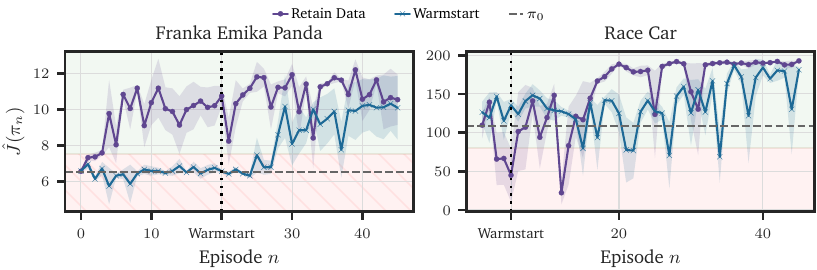}
    \caption{Online performance when retaining data used for training in simulation. Simulation data stabilizes learning even without priming the online buffer with $\pi_0$.}
    \label{fig:sim-data-retention}
\end{figure}

\paragraph{Retaining simulation data.}
We investigate the effect of retaining data collected during simulation on stability and efficiency in online learning, and compare our results with the warmstart setting of \citet{zhou2025efficient}, which assumes that no offline data can be retained during online learning.
Specifically, we load the replay buffer used to train the prior policy $\pi_0$ in simulation, initialize $\alpha = 0.5$ and linearly anneal it to $\alpha = 1$ over five episodes, such that only online data is used thereafter.
The results, shown in \Cref{fig:sim-data-retention}, demonstrate that retaining simulation data substantially improves both learning efficiency and stability.
This result is expected, since the simulation data acts as a regularizer, biasing minibatches used in the  action-value update (\Cref{eq:policy-evaluation}) towards samples with lower approximation error. This dampens sharp distribution shifts during learning, thereby stabilizing learning. As $\alpha$ gradually increases to $1$, only real-world data is used in training, ensuring that optimal performance is ultimately obtained on the real robot.

\section{Franka Emika Panda}
\label{sec:franka-details}

\paragraph{Task.}
We build our simulated and real environments based on the \texttt{PandaPickCubeCartesian} task of \citet{zakka2025mujoco}. The agent observes a $64 \times 64$ grayscale image, together with the end-effector position $(x, y, z)$ in the world frame and the gripper opening. To emulate photometric variability, image brightness is randomly scaled by a uniformly sampled factor at each episode reset, producing observations that vary in illumination and contrast.
The agent acts in Cartesian space through a continuous four-dimensional action vector
\begin{equation*}
  a = (\Delta x, \Delta y, \Delta z, g),
\end{equation*}
where $(\Delta x, \Delta y, \Delta z)$ represent incremental translational displacements of the gripper and $g$ controls the opening and closing of the parallel fingers. The reward function in our setup follows the progress-based reward function of \citet{zakka2025mujoco}, which encourages the agent to make incremental improvements towards the goal within episodes: approach the cube, lift it and move it towards the goal position until reaching it.

\paragraph{Success Criterion.}
An episode is considered successful when the Euclidean distance between the cube and the designated target position falls below a threshold of $0.05\,\mathrm{m}$. Episodes terminate early upon success or if the cube falls off the workspace.

\begin{figure}
    \centering
    \includegraphics[width=\linewidth, trim={0 1.6cm 0 0}, clip]{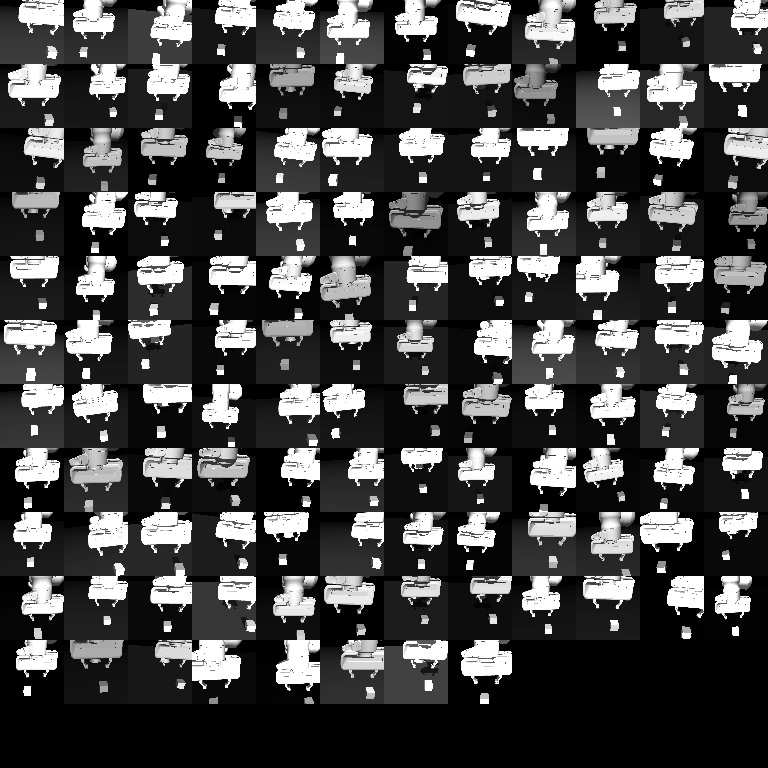}
    \caption{Domain-randomized environments in the simulated Franka Emika Panda robot.}
    \label{fig:dr-franka-emika-panda}
\end{figure}

\paragraph{Domain Randomization.}
We follow the same domain randomization scheme as \citet{zakka2025mujoco}. We provide its details here for completeness.
The randomized parameters include:
\begin{itemize}
    \item \textbf{Lighting:} Randomized light position, orientation, and whether shadows are cast.
    \item \textbf{Camera Pose:} Small perturbations to camera position and orientation.
    \item \textbf{Material and Color:} Randomization of the floor color (grayscale shades) and geometric material identifiers for scene objects.
    \item \textbf{Scene Brightness:} Multiplicative scaling of rendered image intensity to vary illumination.
\end{itemize}
These variations, shown in \Cref{fig:dr-franka-emika-panda}, are independently sampled for each parallel simulation instance, ensuring that the learned policy encounters a diverse range of visual and geometric conditions during training.

\section{Unitree Go1}
\label{sec:go1-details}
\begin{figure}
    \centering
    \begin{subfigure}[b]{\textwidth}
    \centering
    \begin{tabular}{c @{\hskip 5pt} c @{\hskip 5pt} c @{\hskip 5pt} c @{\hskip 5pt} c @{\hskip 5pt} c @{\hskip 5pt} c}
    \rotatebox{90}{$\pi_0$} &
        \begin{tikzpicture}
            \node[anchor=south west, inner sep=0] (image1) at (0, 0) 
                {\includegraphics[width=0.15\textwidth, clip]{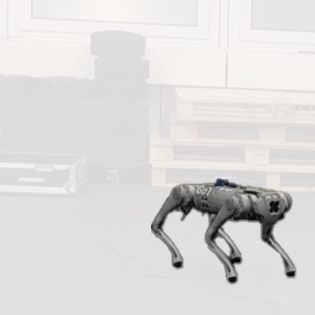}};
        \end{tikzpicture} &
        \begin{tikzpicture}
            \node[anchor=south west, inner sep=0] (image2) at (0, 0) 
                {\includegraphics[width=0.15\textwidth, clip]{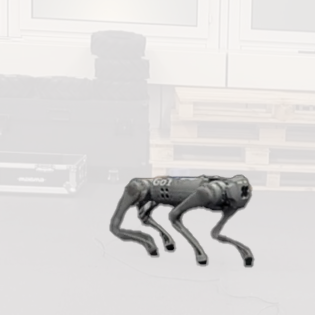}};
        \end{tikzpicture} &
        \begin{tikzpicture}
            \node[anchor=south west, inner sep=0] (image3) at (0, 0) 
                {\includegraphics[width=0.15\textwidth, clip]{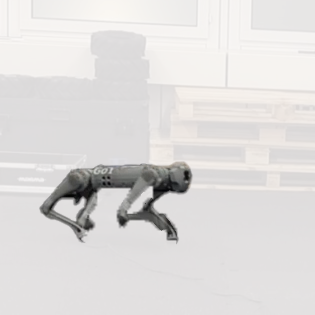}};
        \end{tikzpicture} &
        \begin{tikzpicture}
            \node[anchor=south west, inner sep=0] (image4) at (0, 0) 
                {\includegraphics[width=0.15\textwidth, clip]{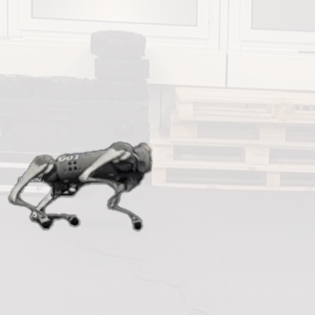}};
        \end{tikzpicture} &
        \begin{tikzpicture}
            \node[anchor=south west, inner sep=0] (image5) at (0, 0) 
                {\includegraphics[width=0.15\textwidth, clip]{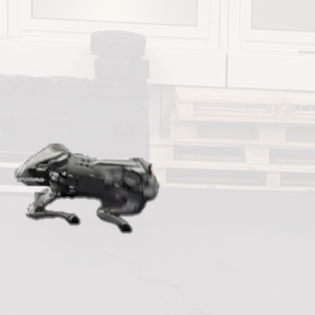}};
        \end{tikzpicture} &
        \begin{tikzpicture}
            \node[anchor=south west, inner sep=0] (image6) at (0, 0) 
                {\includegraphics[width=0.15\textwidth, clip]{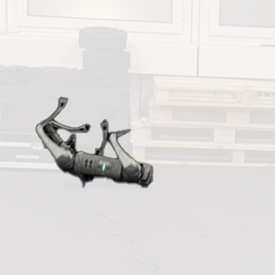}};
        \end{tikzpicture} \\
    \rotatebox{90}{$\pi_N$} &
        \begin{tikzpicture}
            \node[anchor=south west, inner sep=0] (image7) at (0, 0) 
                {\includegraphics[width=0.15\textwidth, clip]{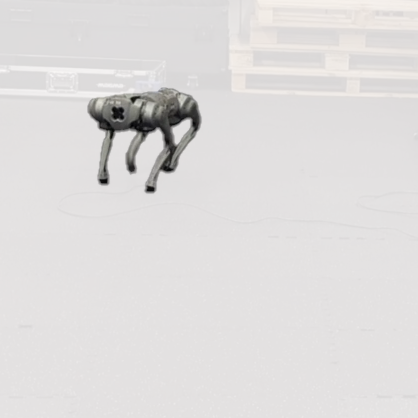}};
        \end{tikzpicture} &
        \begin{tikzpicture}
            \node[anchor=south west, inner sep=0] (image8) at (0, 0) 
                {\includegraphics[width=0.15\textwidth, clip]{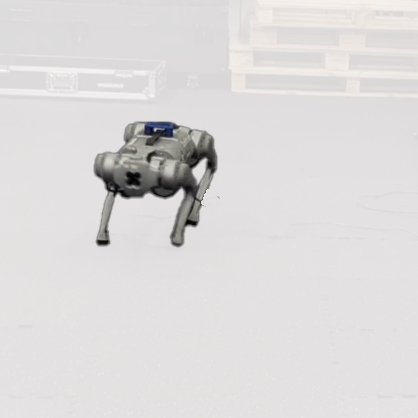}};
        \end{tikzpicture} &
        \begin{tikzpicture}
            \node[anchor=south west, inner sep=0] (image9) at (0, 0) 
                {\includegraphics[width=0.15\textwidth, clip]{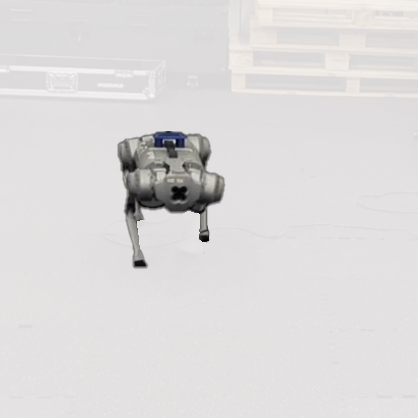}};
        \end{tikzpicture} &
        \begin{tikzpicture}
            \node[anchor=south west, inner sep=0] (image10) at (0, 0) 
                {\includegraphics[width=0.15\textwidth, clip]{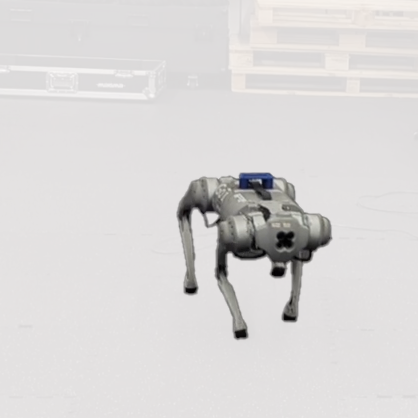}};
        \end{tikzpicture} &
        \begin{tikzpicture}
            \node[anchor=south west, inner sep=0] (image11) at (0, 0) 
                {\includegraphics[width=0.15\textwidth, clip]{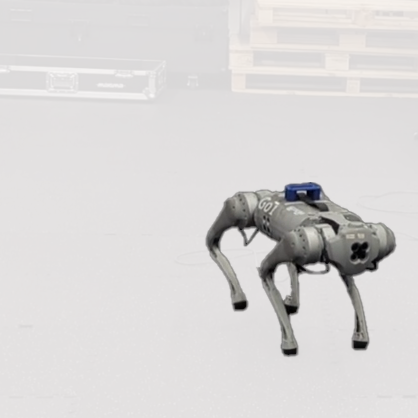}};
        \end{tikzpicture} &
        \begin{tikzpicture}
            \node[anchor=south west, inner sep=0] (image12) at (0, 0) 
                {\includegraphics[width=0.15\textwidth, clip]{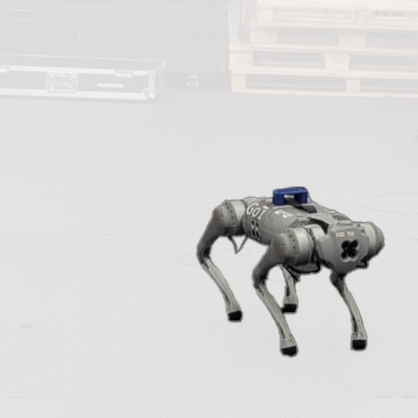}};
        \end{tikzpicture} \\
    \end{tabular}
    \end{subfigure}
    \caption{Robustness to commands that were not seen during training in simulation improves over training episodes.}
    \label{fig:go1-learning-demo}
\end{figure}
\looseness=-1
We use the \texttt{FlatTerrainGo1Joystick} environment from MuJoCo Playground \citep{zakka2025mujoco} in our experiments. Our real-world environment matches the simulated task in all aspects.
Our main deviation from \citet{zakka2025mujoco} is in the linear and angular velocity commands we sample. Specifically, \citet{zakka2025mujoco} samples uniformly from the range~$[ \pm 1.5, \pm 0.8, \pm 1.2]$ during training in simulation, allowing for very good transfer to the real robot since this range of velocities is rather large. We instead train in simulation using the range~$[\pm 0.5, \pm 0.8, \pm 1.2]$, which we also use when deploying to real. Since this range is relatively smaller than the one used by \citet{zakka2025mujoco}, transfer to the real robot is much more challenging because the resulting policy from simulation is less robust when deployed on the real robot. We demonstrate that in \Cref{fig:go1-learning-demo} where we show a trajectory of the prior policy $\pi_0$ and a trajectory after training with improved stability.

\section{Race Car}
\label{sec:rccar-details}
\begin{figure}[ht]
    \centering
    % Timeline arrow
    % Image grid
    \begin{subfigure}[b]{\textwidth}
    \centering
    \begin{tabular}{c @{\hskip 5pt} c @{\hskip 5pt} c @{\hskip 5pt} c @{\hskip 5pt} c}
    \rotatebox{90}{$\pi_0$} &
        \begin{tikzpicture}
            \node[anchor=south west, inner sep=0] at (0, 0)
                {\includegraphics[width=0.23\textwidth, clip]{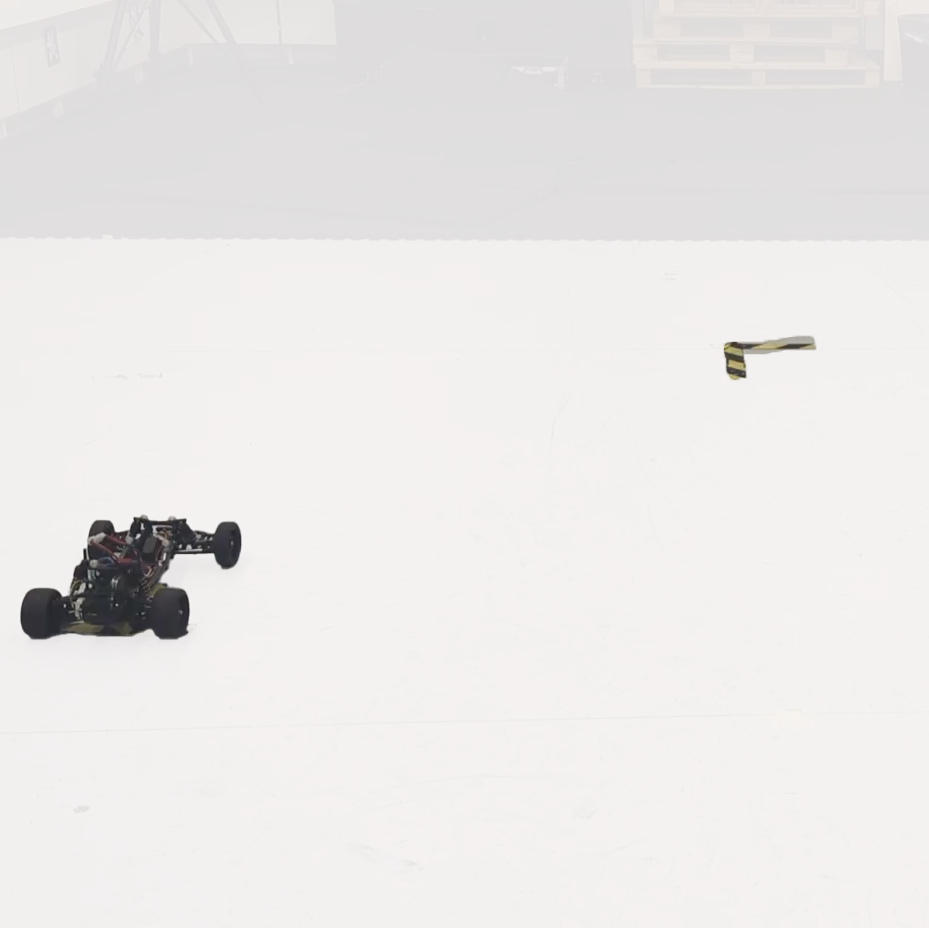}};
        \end{tikzpicture} &
        \begin{tikzpicture}
            \node[anchor=south west, inner sep=0] at (0, 0)
                {\includegraphics[width=0.23\textwidth, clip]{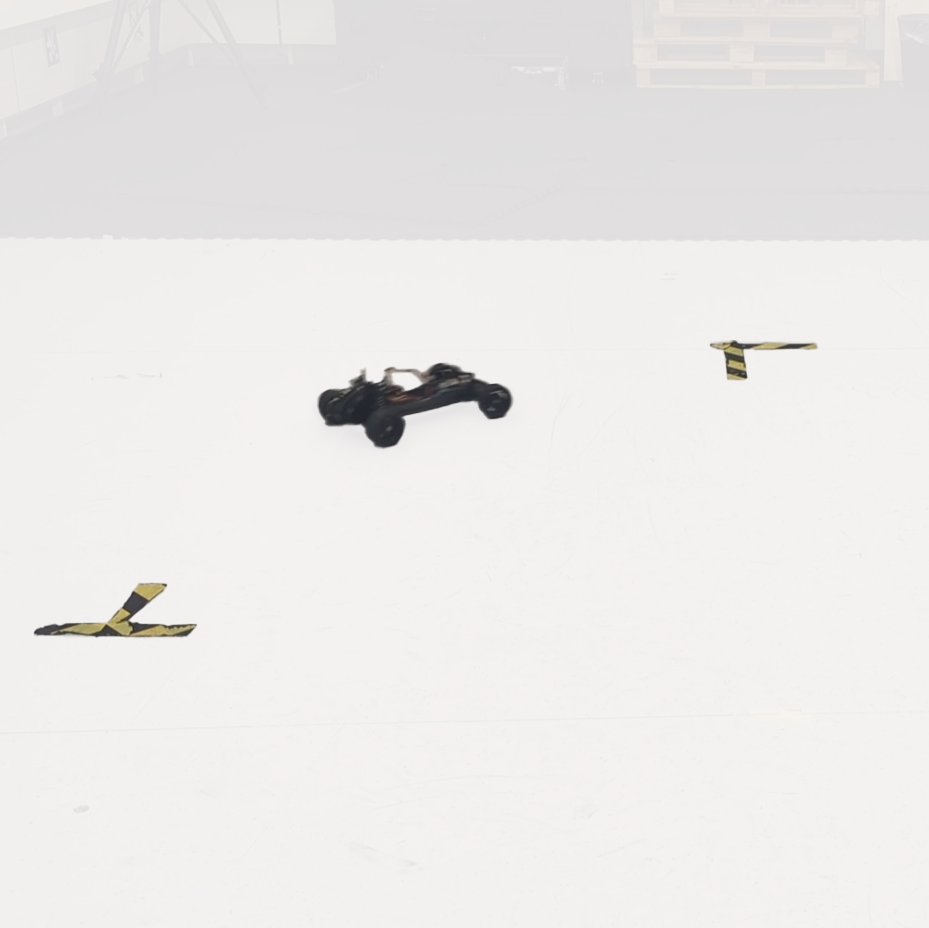}};
        \end{tikzpicture} &
        \begin{tikzpicture}
            \node[anchor=south west, inner sep=0] at (0, 0)
                {\includegraphics[width=0.23\textwidth, clip]{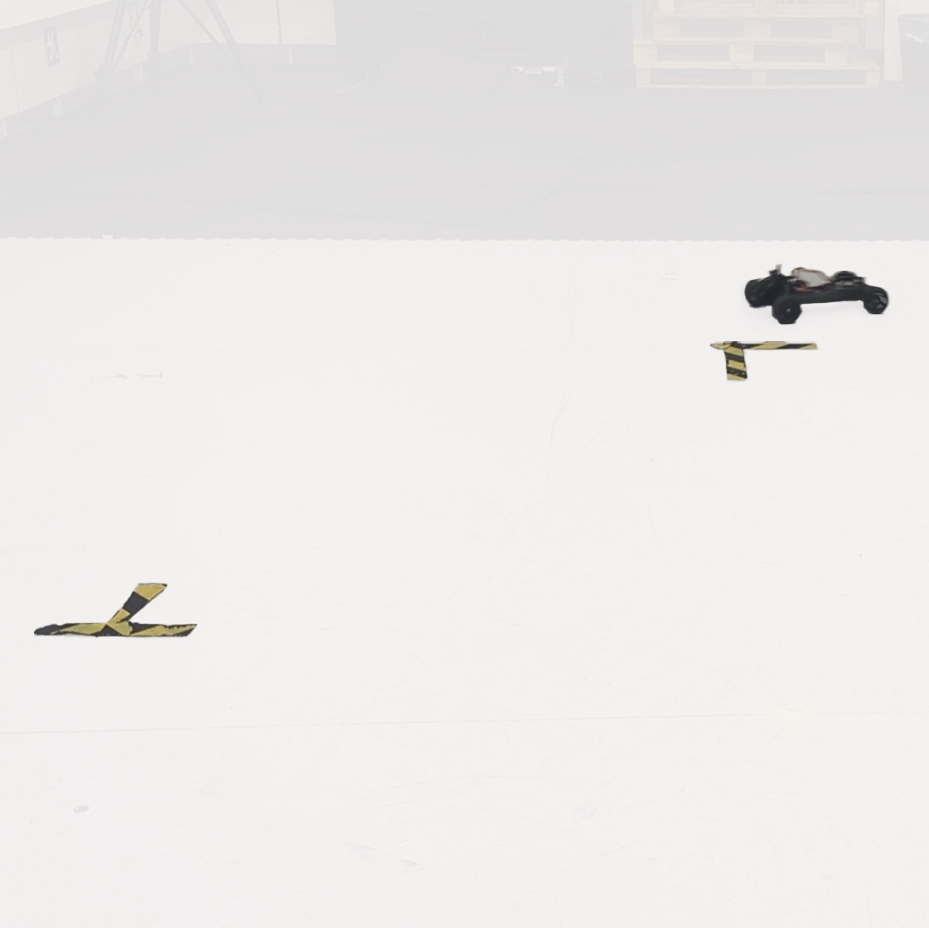}};
        \end{tikzpicture} &
        \begin{tikzpicture}
            \node[anchor=south west, inner sep=0] at (0, 0)
                {\includegraphics[width=0.23\textwidth, clip]{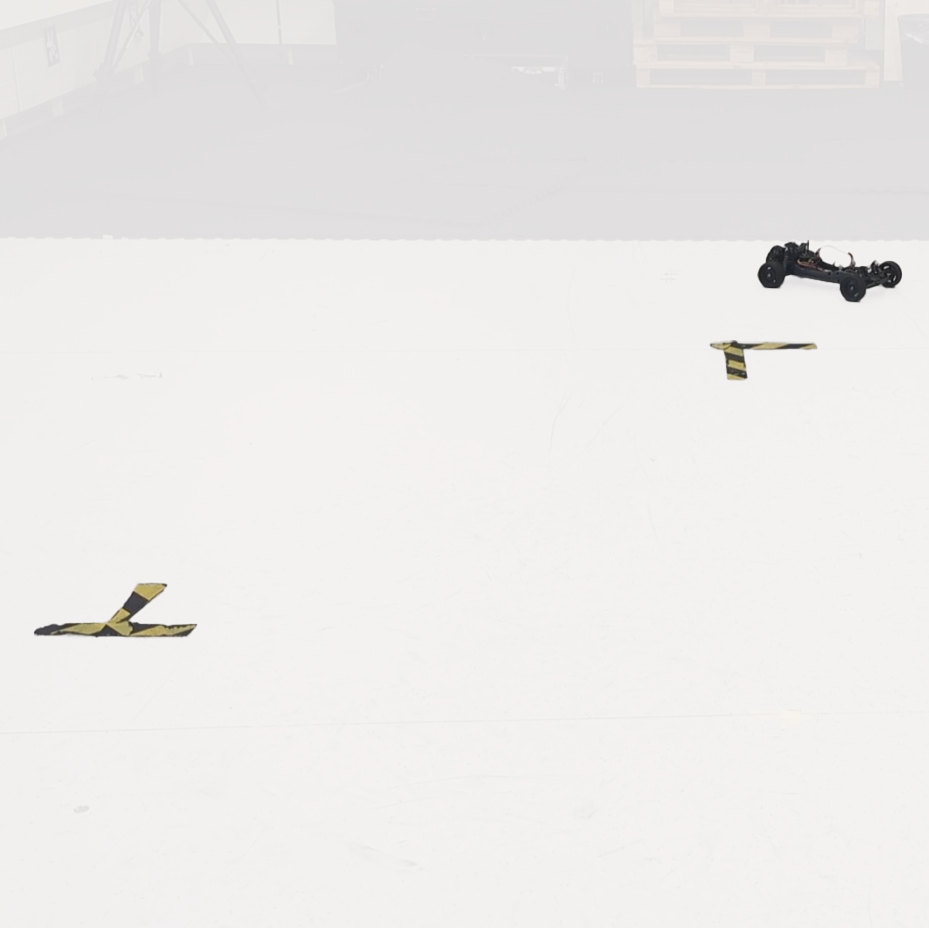}};
        \end{tikzpicture} \\
    \rotatebox{90}{$\pi_N$} &
        \begin{tikzpicture}
            \node[anchor=south west, inner sep=0] at (0, 0)
                {\includegraphics[width=0.23\textwidth, clip]{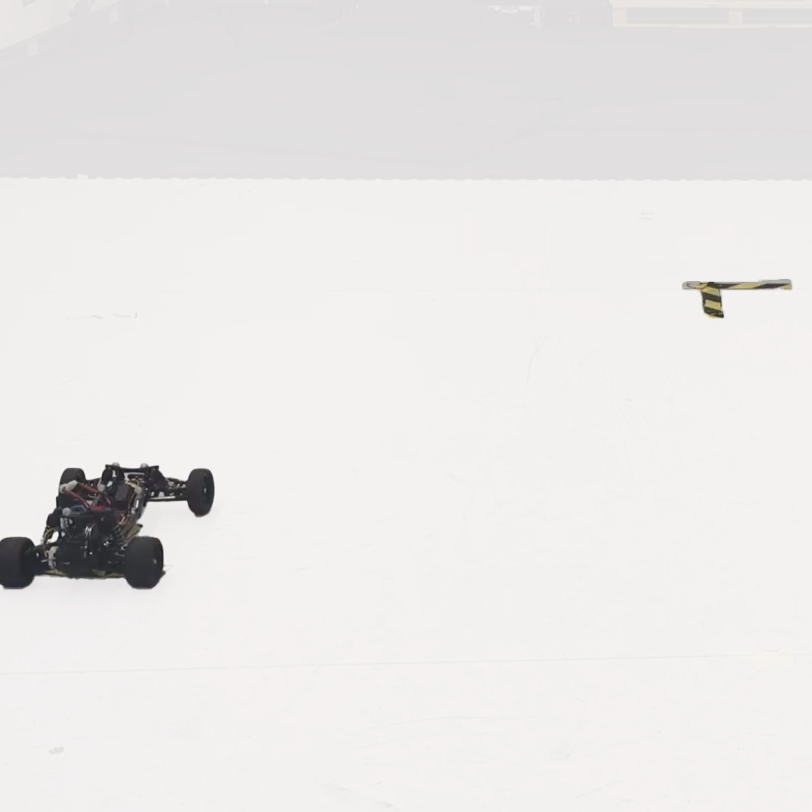}};
        \end{tikzpicture} &
        \begin{tikzpicture}
            \node[anchor=south west, inner sep=0] at (0, 0)
                {\includegraphics[width=0.23\textwidth, clip]{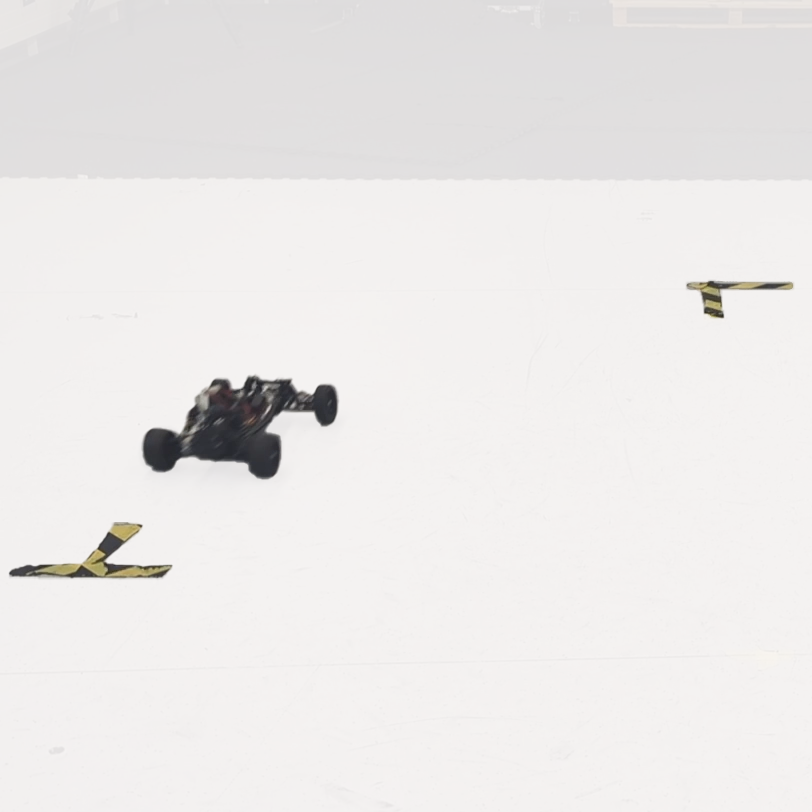}};
        \end{tikzpicture} &
        \begin{tikzpicture}
            \node[anchor=south west, inner sep=0] at (0, 0)
                {\includegraphics[width=0.23\textwidth, clip]{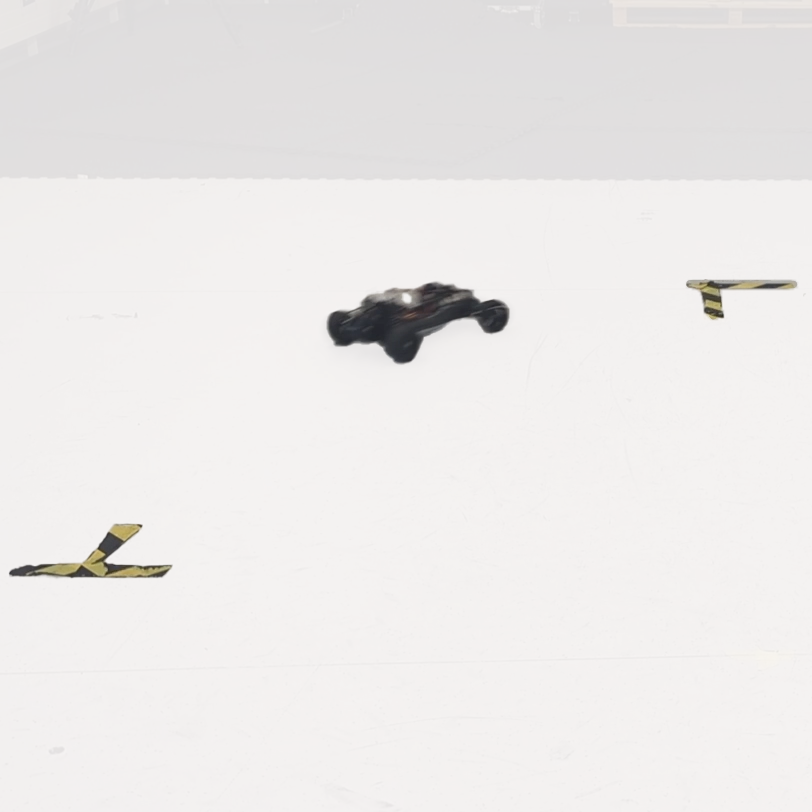}};
        \end{tikzpicture} &
        \begin{tikzpicture}
            \node[anchor=south west, inner sep=0] at (0, 0)
                {\includegraphics[width=0.23\textwidth, clip]{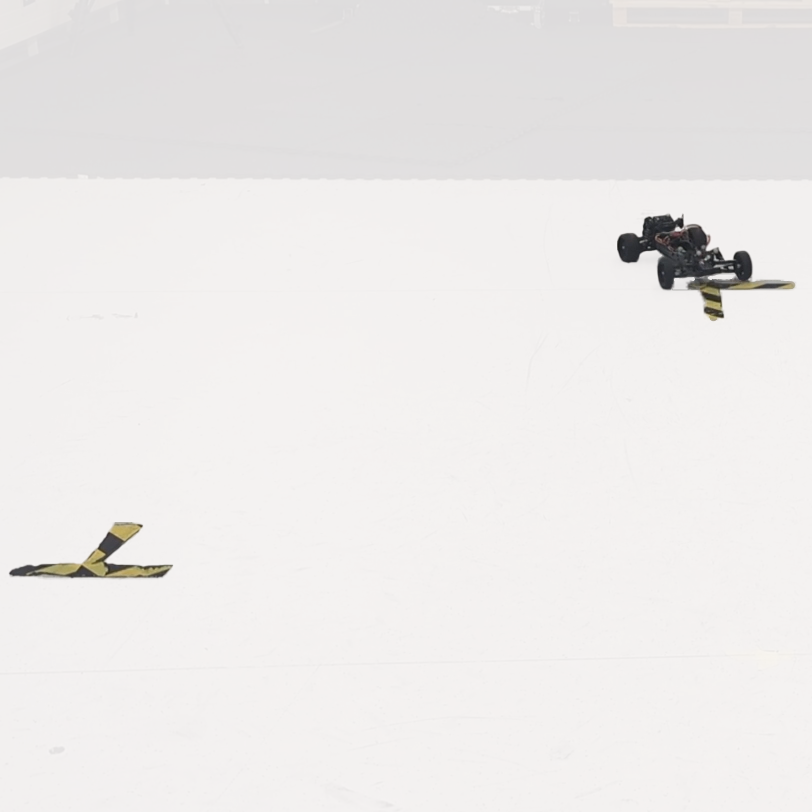}};
        \end{tikzpicture} \\
    \end{tabular}
    \end{subfigure}

    \caption{The prior policy $\pi_0$ overshoots the goal, obtaining low rewards due to the sparse reward in this task. After $20$ trials, the finetuned policy $\pi_N$ reaches the goal faster and with greater precision.}
    \label{fig:rccar-learning-demo}
\end{figure}

\widowpenalty=10000
\paragraph{Task.}
We simulate car dynamics following \citet{kabzan2020amz}. Their model captures the car's motor and tire dynamics, however due to the sim-to-real gap, modeling car drifts is rather inaccurate and leads to the car overshooting the goal position, shown in \Cref{fig:rccar-learning-demo}. 
At each timestep, the agent observes the full vehicle state (2D positions and velocities) and outputs a continuous $2$D action (steering, throttle). The reward is defined as
\begin{equation*}
    r_t(s_t, a_t) \eqdef d_{t-1} - d_t + \1[d_t \leq \epsilon] - \lambda_c \|a_t\|_2 - \lambda_l \|a_t - a_{t-1}\|_2^2,
\end{equation*}
where $d_t = \lVert \mathbf{x}_t - \mathbf{x}_{\text{goal}} \rVert_2$ denotes the Euclidean distance to the goal and $a_t \in \mathbb{R}^2$ is the applied action. The indicator term adds a bonus when within $\epsilon = 0.3$ meters of the goal. $\lambda_c$ penalizes control effort and $\lambda_l$ penalizes action changes.

\section{Implementation Details}
\label{sec:implementation-details}

\paragraph{Hyperparameters.}
\looseness=-1Unless otherwise specified, we use a learning rate of $10^{-5}$ for the actor, and update the actor once every 20 critic updates.
In addition, we use $1250$ updates per episode for all robots, leading to $\eta = 5$ for the Franka Emika Panda and Race Car robots and $\eta \approx 1$ for the Unitree Go1. In principle, higher UTD can further improve sample efficiency, however we opt for a relatively conservative UTD setting when running SAC on real hardware to maintain stable learning for all robots.
All remaining hyperparameters and sweep settings follow the reference open-source implementation in \url{https://github.com/yardenas/panda-rl-kit}.

\paragraph{Synchronous updates.}
Standard off-policy algorithms are typically implemented such that actor-critic updates occur after every real-world transition. 
This approach is challenging in practice, since gradient computations are typically slower than real-time control cycles, especially when increasing the UTD. While synchronous updates after each transition are sometimes feasible \citep[e.g.][]{smith2022walk}, they become impractical for high-frequency control or large models. This suggests a \emph{batch-like} scheme that approaches (iterated) offline RL as $T$ grows. In this scheme, learning occurs \emph{asynchronously} and \emph{episodically} every $T$ steps, decoupling data collection from optimization. Specific details can be found in our open-source implementation: \url{https://github.com/yardenas/panda-rl-kit}.

\end{document}